\pdfoutput=1
\documentclass[11pt]{article}

\usepackage[]{acl}

\usepackage{times}
\usepackage{latexsym}

\usepackage[T1]{fontenc}
\usepackage[utf8]{inputenc}

\usepackage{microtype}

\usepackage{inconsolata}

\usepackage[whole]{bxcjkjatype}

\usepackage{subcaption}
\usepackage{fontawesome5}

\usepackage{booktabs}
\usepackage{tikz}
\usepackage{graphicx}
\usepackage{multirow}
\usepackage{makecell}

\usepackage{amsmath,amssymb,mathtools}

\DeclareMathOperator*{\argmin}{arg\,min}
\usepackage{mathtools}
\newcommand{\mathboldface}[1]{\boldsymbol{#1}}
\newcommand{\bm}[1]{\mathboldface{#1}}
\usepackage{bbm}

\usepackage[utf8]{inputenc}
\usepackage{soulutf8}

\usepackage{todonotes}

\usepackage{hyperref}
\usepackage{breakurl}
\usepackage{url}

\usepackage{enumerate}
\usepackage{enumitem}

\usepackage{xcolor}

\definecolor{purple}{HTML}{7030A0} 
\definecolor{orange}{HTML}{FF3E00} 

\usepackage{amsmath}
\usepackage{amssymb}
\usepackage{amsfonts}
\usepackage{enumitem}

\usepackage{tikz}
\newcommand{\ColoredText}[2]{%
  \tikz[baseline=(X.base)] 
    \node[rectangle, rounded corners, fill=#1, text=black, inner sep=2pt] (X) {#2};%
}

\usepackage{CJKutf8}

\newcommand{\Chinese}[1]{{\begin{CJK*}{UTF8}{gbsn}#1\end{CJK*}}}

\title{On Entity Identification in Language Models}

\author{
Masaki\,Sakata$^{1,2}$\hspace{1em}
Benjamin\,Heinzerling$^{2,1}$\hspace{1em}
Sho\,Yokoi$^{3,1,2}$ \\ 
\textbf{Takumi\,Ito}$^{1,4}$\hspace{1em}
\textbf{Kentaro\,Inui}$^{5,1,2}$\\[2pt]
$^{1}$ Tohoku University\hspace{1em}
$^{2}$ RIKEN\hspace{1em}
$^{3}$ NINJAL\hspace{1em}
$^{4}$ Langsmith Inc.\hspace{1em}
$^{5}$ MBZUAI\hspace{1em}
\\
\texttt{sakata.masaki.s5@dc.tohoku.ac.jp}\hspace{1em}
\texttt{benjamin.heinzerling@riken.jp} \\
\texttt{yokoi@ninjal.ac.jp} \hspace{1em}
\texttt{t-ito@tohoku.ac.jp} \hspace{1em}
\texttt{kentaro.inui@mbzuai.ac.ae} \\
}

\begin{document}
\maketitle

\begin{abstract}
We analyze the extent to which internal representations of language models (LMs) identify and distinguish mentions of named entities, focusing on the many-to-many correspondence between entities and their mentions.
We first formulate two problems of entity mentions --- ambiguity and variability --- and propose a framework analogous to clustering quality metrics. 
Specifically, we quantify through cluster analysis of LM internal representations the extent to which mentions of the same entity cluster together and mentions of different entities remain separated.
Our experiments examine five Transformer-based autoregressive models, showing that they effectively identify and distinguish entities with metrics analogous to precision and recall ranging from 0.66 to 0.9.
Further analysis reveals that entity-related information is compactly represented in a low-dimensional linear subspace at early LM layers.
Additionally, we clarify how the characteristics of entity representations influence word prediction performance.
These findings are interpreted through the lens of isomorphism between LM representations and entity-centric knowledge structures in the real world, providing insights into how LMs internally organize and use entity information.
\faGithub: \href{https://github.com/masaki-sakata/entity-identification/tree/main}{https://github.com/masaki-sakata/entity-identification}
\end{abstract}

\section{Introduction}
\label{sec:introduction}

\begin{figure}[ht]
\centering
\includegraphics[width=\columnwidth]{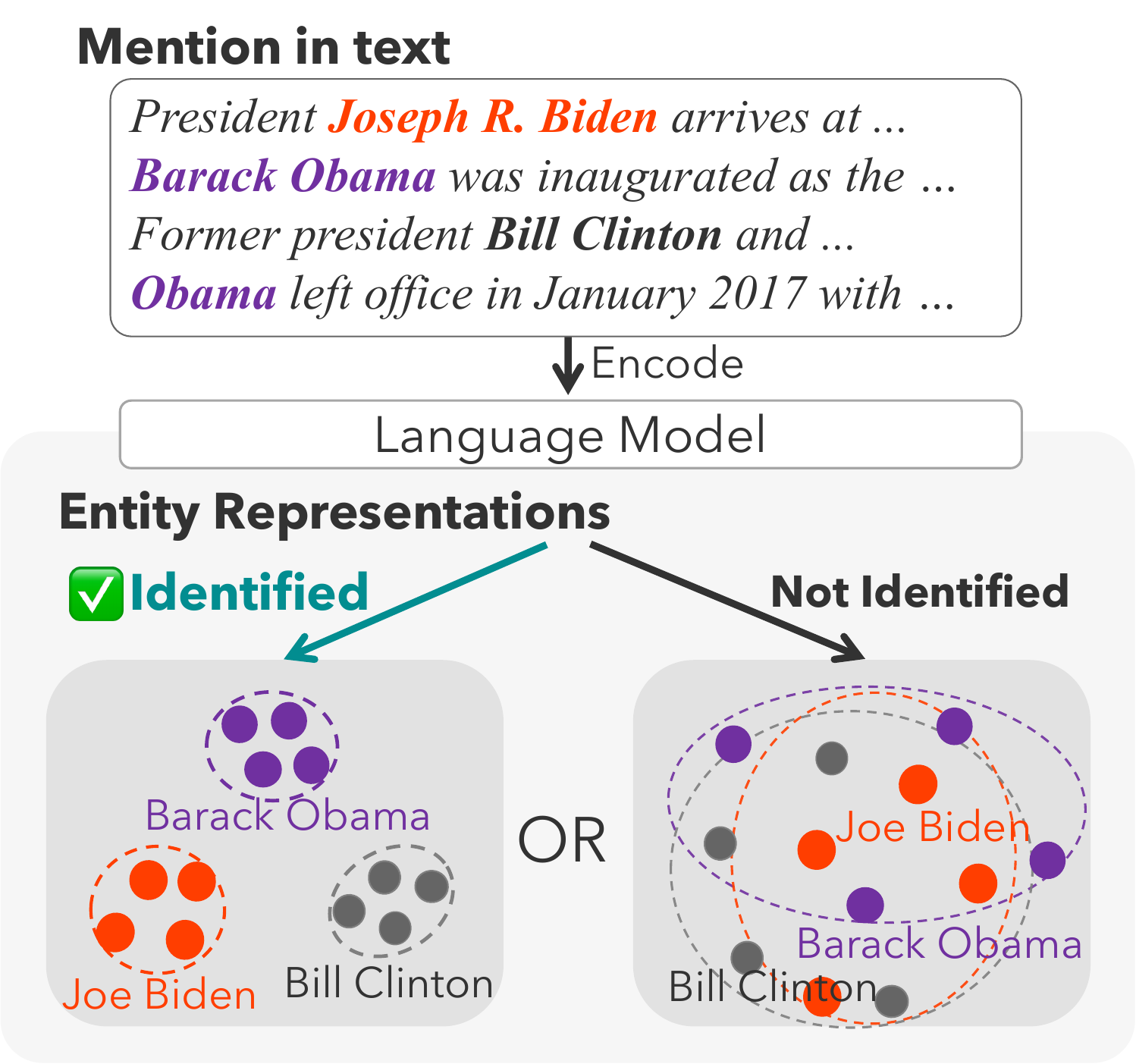}
\caption{
Illustration of \emph{entity identification}.
If ``\textcolor{purple}{Barack Obama}'' or ``\textcolor{orange}{Joe Biden}'' are represented in various mentions and contexts but still form a cluster as a single entity in the representation space, then we observe successful entity identification.
% (as shown on the left side).
Such entity identification in the representation space suggests that the LM's representations reflect certain aspects of the structure of real-world, entity-centric knowledge.
}
\label{fig:pontie}
\end{figure}

Transformer-based language models (LMs) have been demonstrated to be able to recall factual knowledge composed of entities and relations~\citep{petroni-etal-2019-language, jiang-etal-2020-know, heinzerling-inui-2021-language,cohen-etal-2023-crawling}. For example, when given the query ``Barack Obama was born in \_\_'', LMs can predict appropriate words like `Hawaii''. To understand how LM predicts such entities (e.g. ``Hawaii''), extensive internal analyzes have been performed from an information flow perspective~\citep{dai-etal-2022-knowledge, meng2022locating, geva-etal-2023-dissecting}.

\begin{figure}[t]
\centering
\includegraphics[width=\linewidth]{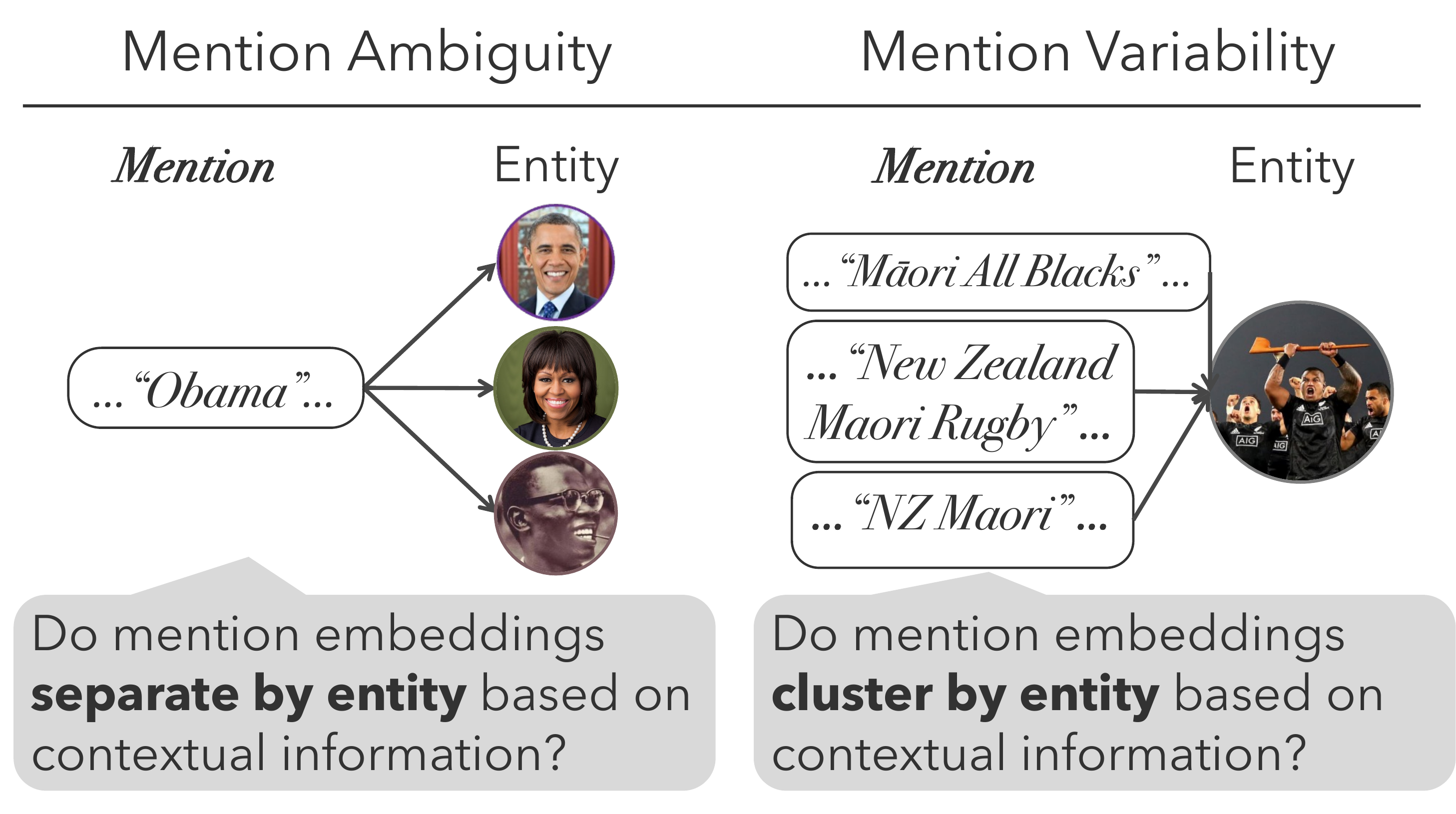}
\caption{
Two major factors that can make entity identification difficult are ambiguity and variability of entity mentions.
}
\label{fig:factors_pontie}
\end{figure}

However, while previous work has shed light on the internal mechanisms of LMs~\citep{dai-etal-2022-knowledge, meng2022locating, geva-etal-2023-dissecting}, our understanding remains limited in scope.
For instance, when processing mentions like ``Obama'' or ``Barack Obama'', it remains unclear whether LMs encode them as the same individual based on context~(Figure~\ref{fig:pontie}).
Although encoding mentions as distinct entities may appear straightforward, there are two major real-world challenges~(Figure~\ref{fig:factors_pontie}).
One is \textbf{mention ambiguity}: a single mention could potentially refer to multiple different entities. 
The other is \textbf{mention variability}: multiple mentions with different surface forms exist for a single entity.
Notably, previous studies have not addressed queries containing mention ambiguity or mention variability, leaving a gap in our understanding of how LMs handle these fundamental challenges.
Properly resolving mention ambiguity and variability is crucial to language understanding.

We assume that encoding where mentions are distinguished by their corresponding entities, as shown in Figure~\ref{fig:pontie} (left), is an appropriate representation and investigate to what extent LMs distinguish entities in their internal representations. 
We call the process of distinguishing entities in internal representations ``entity identification''.
To evaluate entity identification, we used Purity~\citep{Zhao2001CriterionFF} and Inverse Purity (IP).
This approach allows us to directly measure whether the model's representations for the same entity form compact clusters that differ from other entities.

Our experiments with five autoregressive LMs demonstrated that they achieve an AUC of approximately 0.8--0.9 for mention ambiguity (\S\ref{subsec:Mention ambiguity}) and 0.66--0.8 even in the presence of mention variability (\S\ref{subsec:Mention variability}).
LM scores outperformed the baselines, demonstrating their ability to identify entities.
We also find that entity information forms low-dimensional subspaces in early layers (\S\ref{subsec:representation_analysis}), with better entity representation separation improving word prediction (\S\ref{subsec:entity_representation_and_word_pred}).
Finally, we discuss these findings through the lens of isomorphism between LM representations and real-world entity-centric knowledge structures (\S\ref{sec:discussion}). 
This provides evidence that LMs encode discrete structural knowledge through text-only training, extending previous observations~\citep{abdou-etal-2021-language, chen2023correlationlargelanguagemodels, gurnee2024language, park2024the} toward a systematic understanding of how LM representation spaces correspond to structural knowledge.

%%%%%%%%%%%%%%%%%%%%%%%%%%%%%%%%%%%

\section{Entity Identification} 
\label{sec:method}
Simply put, our goal is to find out if LMs ``know''\footnote{This loose, anthropomorphizing use of the word ``know'' will be made more precise in the following section.} that different mentions of the same real-world entity refer to the same entity and, conversely, that mentions of different entities refer to different entities.
For example, in discussions about ``Barack Obama'' or ``President Obama'', a LM with world knowledge should be able to recognize that both mentions refer to the same person.

\subsection{Why is Entity Identification Difficult?}
\label{sec:why_EI_difficult}
What makes entity identification difficult can be attributed to the ambiguity and variability of a mention (Figure~\ref{fig:factors_pontie}).
From the perspective of mention ambiguity, ``Obama'' could refer to ``Barack Obama'' or ``Michelle Obama'' depending on the context.
From the perspective of mention variability, for example, ``Māori All Blacks'' and ``New Zealand Maori Rugby'' refer to the same rugby team in New Zealand despite using different words.
Resolving mention ambiguity and variability requires effective use of surface and contextual information.

\subsection{Purity and Inverse Purity}
\label{subsec:Purity_IP_F1}
\begin{figure}[t]
\centering
\includegraphics[width=0.9\columnwidth]{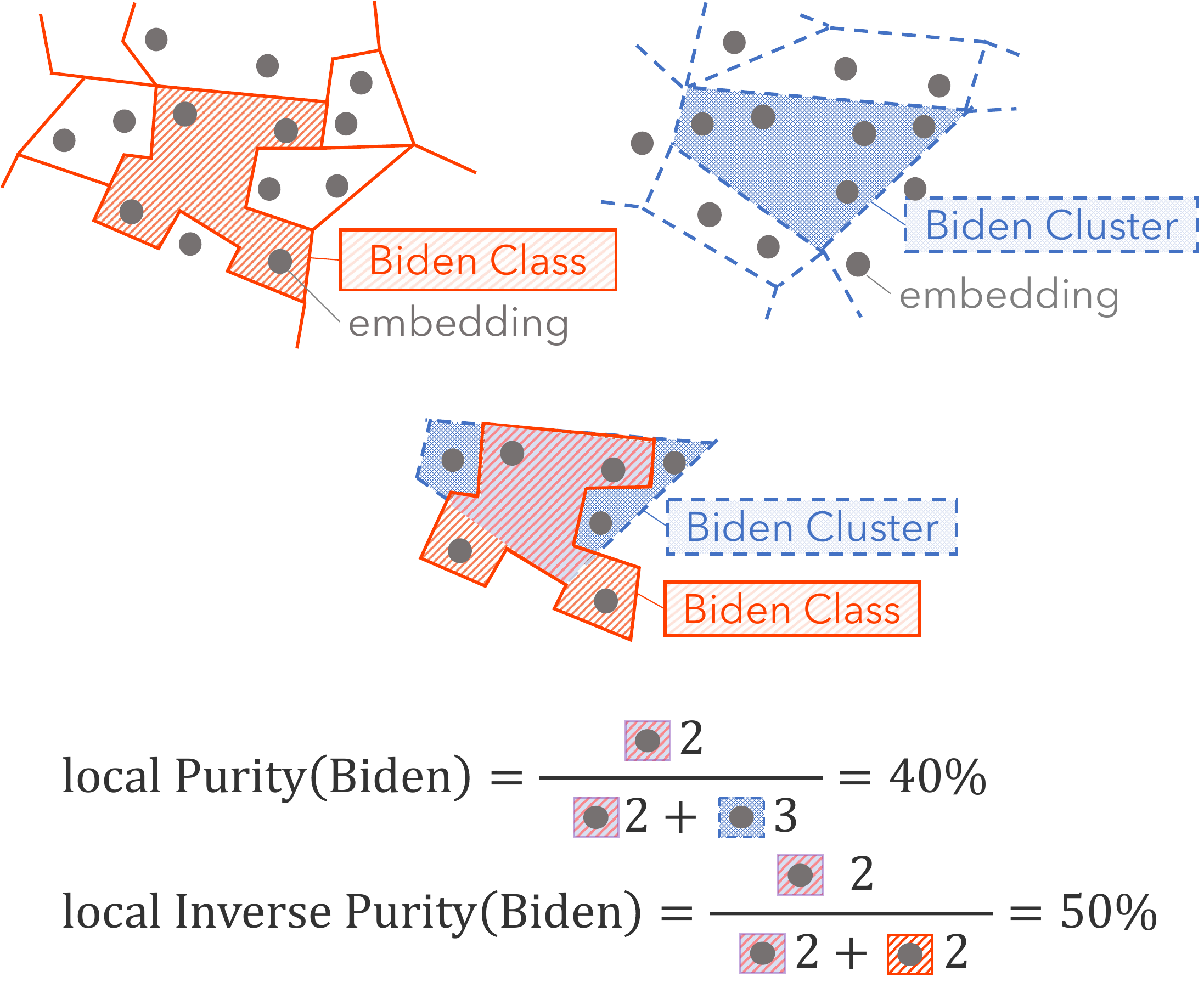}
\caption{
Overview of local Purity and local IP based on Equation~\ref{eq:local_purity} and Equation~\ref{eq:local_inverse_purity}.
As in Equation~\ref{eq:f1}, a high F1 score (purity and IP) indicates not only that embeddings of the same {\color{orange} class} are contained within a {\color{blue} cluster}, but also that embeddings of each {\color{orange} class} are appropriately separated into distinct {\color{blue} cluster}.
The circles represent the embeddings corresponding to each mention.
}
\label{fig:method}
\end{figure}

When a mention with context is input, we aim to quantify the degree to which the LM distinguishes between the entities corresponding to that mention in its representations. 
Intuitively, if the LM effectively distinguishes between entities, the embeddings representing these entities are expected to be clearly separated based on conceptual similarity (as illustrated in the example of ``Barack Obama'' in Figure~\ref{fig:pontie}).
To measure how well the embeddings of mentions are separated and encoded on an entity basis, we use an F1 score composed of (local) \textbf{Purity} \cite{Zhao2001CriterionFF} and (local) \textbf{Inverse Purity (IP)}.\footnote{As an alternative to Purity and IP, the Adjusted Rand Index (ARI) \citep{ARI} can also be employed. ARI was implemented in this study, and the results are reported in Appendix~\ref{sec:appendix_ari}; they show the same overall trend as the Purity/IP scores. Although ARI corrects for chance agreement, it is less suitable for fine-grained, entity-level analysis and is therefore omitted from the main text.}
In other words, we assess the geometric locality of a set of embeddings by examining how much they are mixed with other sets of embeddings.
In Figure~\ref{fig:method}, we illustrate several embeddings, showing both the {\color{orange} class} division corresponding to the Biden entity ({\color{orange} Biden Class}) and the {\color{blue} cluster} division formed based on the centroid of the Biden embeddings ({\color{blue} Biden Cluster}). 
The F1 score using (local) Purity and (local) IP indicates how well the {\color{blue} Biden Cluster} matches the true {\color{orange} Biden Class}.
An F1 score of 1.0 indicates that the Biden embeddings are completely separated from other clusters. 
As the score decreases, it indicates that the embeddings are increasingly mixed with other clusters
The detailed computational procedure for Purity, IP and F1 score can be found in the Appendix \ref{sec:Algorithm}.
Note that since this evaluation method relies on distances between representations, it may be affected by the curse of dimensionality, which we examine in \S~\ref{subsec:dimensional_impact}.

The F1 score based on Purity and IP can be considered as one of the methodological approaches to measure how well LMs capture the structure of the world (see Table~\ref{tab:method_comparison} in Appendix).
There is widespread interest in understanding how well LMs capture the structure of the world. 
Typically, the \emph{degree} of correspondence between the world's inherent structure (such as the similarity between colors in color space) and their representation within LMs (such as the similarity between color names in the LM's latent space) is evaluated through methods like representational similarity analysis (RSA)~\citep{abdou-etal-2021-language, patel2022mapping, chen2023correlationlargelanguagemodels, gurnee2024language, DBLP:conf/iclr/HernandezSHMWAB24, park2024the}.
For entities, however, the crucial aspect is whether pairs of mentions refer to the same entity or different ones. 
Our proposed clustering-based method measures the \emph{degree} of alignment between this binary structure (``same or different'') in the world and in LMs. 
This approach enables a more natural evaluation of word pairs that, while semantically very similar, would compromise factuality if substituted for one another (such as mentions of entities).

\section{Experimental Setup}
\label{sec:Experiments_setup}
\subsection{Models}
\label{paragh:model} 
We used five variants of LMs, including GPT-2~\citep{gpt2}, Llama-2 (7B, 13B)~\citep{touvron2023llama_2}, Llama-3 (8B)~\citep{dubey2024llama3herdmodels}, and Mistral (7B)~\citep{jiang2023mistral7b}.
The details are shown in Appendix~\ref{sec:appendix_model}, Table~\ref{tab:models_detail}.

We use the following three baselines:
(1) Random embeddings: Individual embeddings randomly sampled from a normal distribution for each mention occurrence. Different embeddings are assigned even for identical mentions.
(2) Unique mention embeddings: Fixed embeddings randomly sampled from a normal distribution for each unique mention. The same embedding is always assigned to identical mentions (e.g., the mention ``Obama'' consistently uses the same embedding).
(3) FastText embeddings\footnote{We use embeddings trained on Wikipedia 2017, the UMBC webbase corpus~\citep{han-etal-2013-umbc}, and the statmt.org news dataset.}~\citep{mikolov-etal-2018-advances}: Pre-trained static word embeddings. When a mention consists of multiple tokens, we use the average embedding of those tokens.
Unlike LMs, unique mention embeddings and FastText embeddings only capture surface-level information of mentions.

\subsection{Entity Representations}
Following the configurations of \citet{abdou-etal-2021-language,bommasani-etal-2020-interpreting,vulic-etal-2020-probing}, we use the hidden states of LMs as representations of a given word. 
The term ``Entity Representations'' refers to the representations that correspond to the mention of an entity. 
In the case of autoregressive LMs, several works have indicated that information related to an entity is concentrated in the last token of a mention~\citep{meng2022locating, geva-etal-2023-dissecting, heinzerling-inui-2024-monotonic}. 
Thus, we report the results using embeddings of the last token.\footnote{The results using average embeddings of subwords are reported in Appendix~\ref{subsec:appendix_mention_amb_ave} and \ref{subsec:appendix_mention_vari_ave}.}
Since autoregressive LMs cannot read the context that appears after analyzing the mention, we repeat the input sentence following \citet{springer2024repetition}.
Specifically, the autoregressive LM reads the entire sentence once and then uses the embedding of the mention appearing the second time as the entity representation.\footnote{
Given a sentence, ``Alice went to Paris,'' we input ``Alice went to Paris. Alice went to Paris'' to the LM and used the embedding of the second ``Alice'' as the entity representation.
}

\subsection{Data}
\label{pargh:data}
We used the ZELDA-TRAIN dataset~\citep{milich-akbik-2023-zelda}, employed for the entity disambiguation task. 
We filter out entities with fewer than five instances, as sparse instances could lead to biased centroid positions when computing the average semantic representation of each entity. 
Additionally, instances where the length of the tokenized sentences exceeded 500 tokens were excluded to ensure compatibility with a wide range of LMs, setting a conservative input length limit.
Table~\ref{tab:data} in Appendix~\ref{subsec:appendix_data} shows the statistical information of the dataset.

\subsection{Preliminary Study: Dimensional Impact}
\label{subsec:dimensional_impact}

\begin{figure}[t]
\centering
\includegraphics[width=\linewidth]{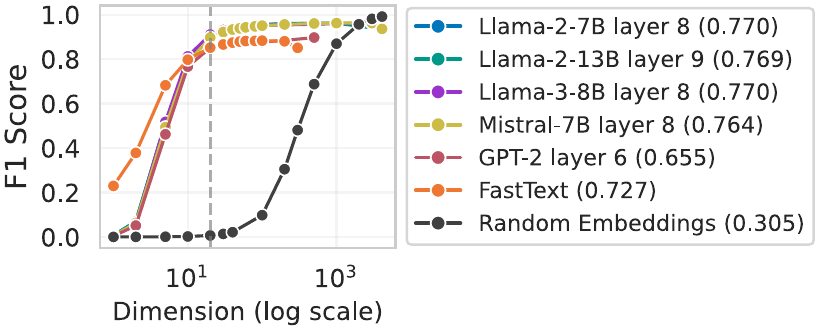}
\caption{
Comparison of F1 scores across different dimensions for LMs, FastText, and random embeddings.
The vertical dashed line indicates dimension=20.
For each model, the layer with the highest AUC score is visualized.
}
\label{fig:dim}
\end{figure}

We first investigate the effect of dimensionality on entity identification. 
This is crucial because our evaluation method focuses on the distance between representations, making it potentially susceptible to the curse of dimensionality.
Using the filtered ZELDA-TRAIN dataset, we compared three kinds of representations: (i) LMs, (ii) FastText, and (iii) random embeddings.
We reduced the LM and FastText representations with linear discriminant analysis (LDA). Random embeddings, in contrast, were generated directly at each target dimensionality.
Figure~\ref{fig:dim} summarizes the results and reveals two main observations.
First, random embeddings score almost zero in low dimensions but nearly saturate in high dimensions.
This indicates that our metric is affected by the curse of dimensionality.
Second, LMs remain strong even in low dimensions; for example, Llama-2-7B attains an F1 of 0.90 in 20 dimensions, only 3\% lower than its score in 4,096 dimensions.
To determine the optimal dimension, we analyze the rate of performance change ($\Delta F1 / \Delta d$) in different dimension windows. 
The slope of the F1 score of LMs rapidly decreases to approximately 20 dimensions, after which it approaches zero ($|\Delta F1 / \Delta d| < \epsilon$ where $\epsilon = 0.005$ for $d > 20$).
Therefore, we conduct subsequent analyses in 20-dimensional space where the curse of dimensionality has minimal impact.
This efficient encoding of entity information in low-dimensional subspaces aligns with the manifold hypothesis of LM representations \citep{cai2021isotropy, cheng-etal-2023-bridging, valeriani2023the}.

\subsection{Definition of Difficulty}

\subsubsection{Mention Ambiguity}
Mention ambiguity quantifies how ambiguously a mention is assigned to different candidate entities based on its frequency distribution between entities.
For example, the mention ``Oxford'' can refer to different entities depending on the context, such as Oxford University or Oxford as a place name, and these entities are referenced with similar frequency. 
When a mention refers to multiple entities with similar probabilities, the ambiguity becomes high in such cases.
We use the entropy $H$ of the candidate mappings from mention to entity as the mention ambiguity: 
$H = - \sum_{i=1}^{N} p_i \log p_i$.
Here, \( p_i \) represents the proportion of times a mention refers to the candidate entity \( i \).\footnote{The detailed definition of mention ambiguity is provided in Appendix~\ref{subsec:appendix_definition_mention_ambiguity}.}
A higher entropy \( H \) indicates greater uncertainty in the candidate entity corresponding to a mention, suggesting more ambiguity (see Table~\ref{tab:factors} in Appendix).
In our experiments, we used a subset of ambiguous mentions, excluding cases where \( H = 0 \).

\subsubsection{Mention Variability}
The variability of the mention refers to the degree to which different surface forms can vary when referring to the same entity.
For example, the mentions ``Māori All Blacks'' and ``New Zealand Maori Rugby'' refer to the same entity, but their surface forms are significantly different.
In such cases, we want to assign a high score to mention the variability.
We use surface-form dissimilarity, which indicates the surface-level variation of multiple mentions for the same entity, as mention variability.
This dissimilarity \( D \) is formulated as:
\begin{align}
D &= \frac{2}{|M_e|(|M_e|-1)} \nonumber \\
&\sum_{i=1}^{|M_e|-1} \sum_{j=i+1}^{|M_e|} \frac{L(m_i, m_j)}{\max(|m_i|, |m_j|)} \nonumber
\end{align}
\( L(m_i, m_j) \) is the Levenshtein distance\footnote{Levenshtein distance is calculated at the character level.}
between mentions \( m_i \) and \( m_j \), and \( |M_e| \) is the total number of mentions for the same entity.\footnote{
The detailed definition of mention variability is provided in Appendix~\ref{subsec:appendix_definition_mention_variability}.
}
A higher dissimilarity \( D \) indicates greater surface-level differences between mentions for the same entity (see Table~\ref{tab:factors} in Appendix).

\section{Results}
\label{sec:result}
\begin{figure*}[t]
\centering
\includegraphics[width=14cm]{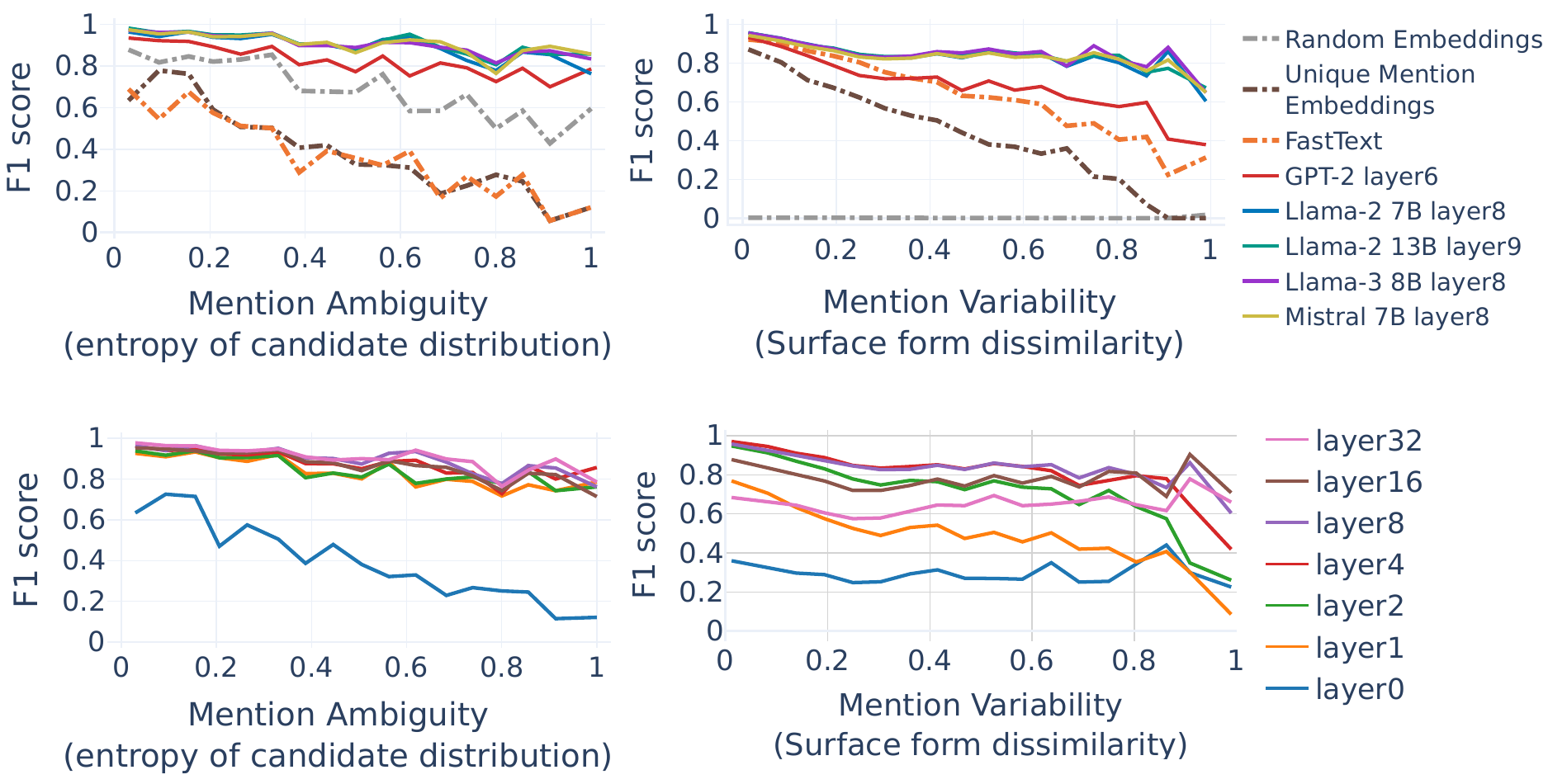}

\caption{Separability (F1 scores) of entity representations for each difficulty~(left: mention ambiguity, right: menti  on variability). 
F1 scores are averaged within each bin on the x-axis. 
Top: Results across all models, using the layer with highest AUC for LM scores. 
Bottom: Layer-wise analysis for Llama-2 7B. 
        Complete results and AUC scores for all models are provided in Appendix \ref{sec:appendix_mention_ambiguity_models} and \ref{sec:appendix_mention_variability_models}.}

\label{fig:result_summary}
\end{figure*}

\subsection{Mention Ambiguity}
\label{subsec:Mention ambiguity}

In the upper-left region of Figure \ref{fig:result_summary} shows the AUC scores of the baseline model and the highest AUC scores achieved across all layers of each LM when controlling for mention ambiguity.
Higher scores on the X-axis indicate more ambiguous mentions, while higher scores on the Y-axis indicate more successful entity identification (entity separated).

LMs achieved AUC scores of 0.8--0.9.
This score is markedly higher than the AUC 
of baselines.\footnote{Random Embeddings generate distinct embeddings for each instance of the same mention, potentially leading to spuriously inflated scores in mention ambiguity. For instance, when the mention ``Obama'' occurs multiple times in the text, each occurrence is assigned a different embedding. While this might result in an embedding space that appears to differentiate between Barack Obama and Michelle Obama, such discrimination is purely coincidental.}
These results demonstrate that LMs distinguish the identity of entities even for ambiguous mentions, compared to baselines.

We found that LMs achieve effective entity identification by stacking layers and contextualizing embeddings.
As shown in the bottom left of Figure~\ref{fig:result_summary}, Llama-2 7B's AUC substantially improved from 0.38 at layer~0 to 0.87 at layer~8.
This transformation can also be qualitatively observed in Appendix Figure~\ref{fig:mention_ambiguous_qualitative_layer-wise_Georgia}.
At layer 0, embeddings of the country name ``Georgia'' and the U.S. state name ``Georgia'' are mixed, but by layer~8, they are clearly separated.
Thus, contextual information is essential for entity identification in cases with mention ambiguity, and LMs can address this challenge through embedding contextualization by stacking layers.

\subsection{Mention Variability}
\label{subsec:Mention variability}

In the upper-right region of Figure \ref{fig:result_summary} shows the AUC scores of the baseline model and the highest AUC scores achieved across all layers of each LM when controlling for mention variability.
Higher scores on the X-axis indicate more variable mentions, while higher scores on the Y-axis indicate more successful entity identification.
The  results showed similar trends to mention ambiguity.\footnote{F1 scores cannot be directly compared between mention ambiguity and mention variability due to different evaluation data units. Mention ambiguity is evaluated on data aggregated by identical mentions, while mention variability is evaluated on an entity basis.}

Overall, LMs achieved AUC scores of 0.66--0.8.
This score stands in stark contrast to the AUC of random embeddings, unique mention embeddings, and FastText.
These results demonstrate that LMs distinguish entity identities even with variable mentions, in contrast to the baselines.
However, all LMs showed a notable decrease in F1 scores when mention variability exceeded 0.8.
For example, the basketball team Saski Baskonia, also known as ``Taugrés,'' is a case with a variability exceeding 0.8.
In such cases, Llama-2 7B recorded an F1 score of 0.5.

We found that LMs achieve effective entity identification by stacking layers and contextualizing embeddings.
As shown in the bottom right of Figure~\ref{fig:result_summary}, Llama-2 7B's AUC substantially improved from 0.3 at layer 0 to 0.81 at layer 8.
Thus, contextual information is essential for entity identification in cases with mention variability, and it was confirmed that LMs can effectively address this challenge through embedding contextualization by stacking layers.
However, it's notable that AUC scores decrease in the later layers (16 and 32) compared to layer 8. 
One possible explanation is that the influence of next-word prediction hinders the process of distinguishing entities in the input representations.

\section{Additional Experiments}
\label{sec:additional_experiment}
We conduct additional experiments to analyze two points: (1) the properties of entity representations (e.g., linear separability) and (2) their impact on word prediction.

\begin{figure}[t]
\centering
\includegraphics[width=\linewidth]{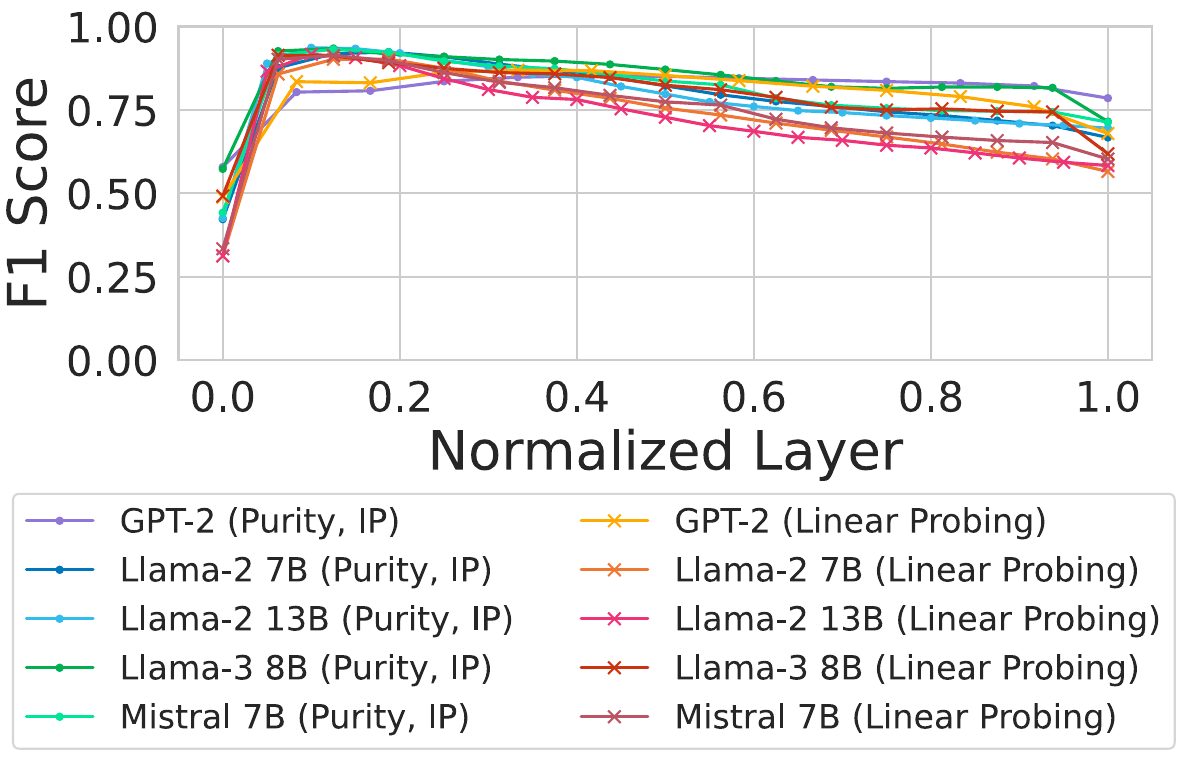}
\caption{
F1 scores based on Purity, IP, and Linear Probing. 
The x-axis represents the relative position of the layers.
All embeddings are reduced to 20 dimensions. 
}
\label{fig:LinearProbing_PurityIP}
\end{figure}

\subsection{Properties of Entity Representations}
\label{subsec:representation_analysis}
\paragraph{Entity Representations are Linearly Separable in Early Layers}
\label{subsubsec:representations_are_linearly}
To determine whether entity representations are encoded in a linearly separable manner, we conduct linear probing. Linear probing serves as an alternative method to Purity and IP, involving training a linear classifier to predict entity labels from representations.
The detailed settings for linear probing are described in Appendix~\ref{subsec:appendix_probe_setting}.
To investigate which layers contain information necessary for entity identification and whether there are differences in trends among Purity, IP, and Linear Probing results, we calculate F1 scores for each layer.
As in \S\ref{sec:result}, the embedding dimensions are reduced to 20 dimensions using LDA.
The results of these analyses are shown in Figure~\ref{fig:LinearProbing_PurityIP}.

Our analysis revealed two main findings.
First, Linear Probing achieved F1 scores of around 0.9.
This suggests that entity representations are encoded in an (almost) linearly separable form.
Previous studies~\citep{olah2020zoom,elhage2022toymodelssuperposition,gurnee2024language} have been accumulating evidence supporting the linear representation hypothesis, which suggests that features within neural networks are represented linearly. 
Our results is further evidence to support this hypothesis.

Second, we found that the maximum F1 score peaks for both Purity, IP, and Linear Probing occur around normalized layer 0.2 (e.g., layers 6-8 in Llama-2 7B). 
This indicates that the information necessary for entity identification is present in the early layers. 
These layer-wise linear probing results align with the findings of \citet{gurnee2024language}, who reported that probing scores for geographical and temporal features peak in early layers. 
While they focused specifically on geographical and temporal features, our analysis extends to a broader range of entities.
Our results also show similarities with \citet{meng2022locating}, who found that entity information is encoded in early layers when processing the last token of entity mentions.
They showed this using activation patching, while we confirmed the same trend through a different approach that examines geometric properties in embedding space.
Considering that our experiments were conducted in a 20-dimensional space, these results suggest that the information necessary for entity identification is encoded in a low-dimensional linear subspace of the early layers in LMs.

\paragraph{Structural Similarity of Entity Representations} 
\label{subsubsec:rsa}
Examining the upper portion of Figure~\ref{fig:result_summary}, we observe similar F1 score patterns across different LMs, particularly among Llama-2, Llama-3, and Mistral. 
This pattern suggests that these models develop comparable structural representations for entities. 
To quantify this observation, we conducted Representation Similarity Analysis (RSA)~\cite{RSA} to measure the structural similarity between representations (see Appendix~\ref{subsec:appendix_rsa} for experimental details).

\begin{figure}[t]
\centering
\includegraphics[width=\linewidth]{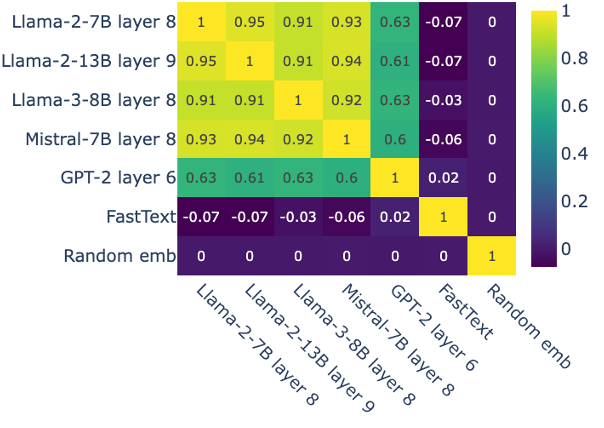}
\caption{
Results of RSA experiments.
Each value represents Spearman's rank correlation coefficient.
LM representations are taken from the layer with the highest F1 score.
Dimensionality is fixed at 20 for all representations.
}
\label{fig:rsa}
\end{figure}

Figure~\ref{fig:rsa} presents our results. 
First, we observe that pairs of LM representations exhibit higher similarity than the static embedding baseline.
Interestingly, Llama-2, Llama-3, and Mistral exhibit stronger inter-model similarity than comparisons involving GPT-2. 
This finding suggests that larger, more capable LMs converge toward similar entity representation structures. 
While our results echo the findings of \citet{The_Platonic_Representation_Hypothesis2024}, we extend their work by focusing specifically on entity representations. 
This structural similarity allows us to interpret LMs through the lens of isomorphism, which we discuss in detail in \S~\ref{sec:discussion}.

\subsection{Effect of Entity Representations Structure on Word Prediction}
\label{subsec:entity_representation_and_word_pred}
To understand how the degree of entity representation separation affects word output, we investigate from two perspectives: output consistency and entity disambiguation accuracy.

\paragraph{Output Consistency}
\label{subsubsec:output_consistency}
\label{subsec:entity_structure_consistency}
\citet{cao-etal-2022-prompt} reported cases where simply changing ``The U.S.'' to ``America'' in the prompt ``The U.S. capital of \_\_'' led to different word prediction results.
In other words, they reported that the variation in surface forms used to refer to the same entity causes inconsistency in LM's word predictions.
In such cases, it would be desirable for the model to consistently predict ``Washington'' for both inputs.
We verify through a simple experiment that this word prediction inconsistency is related to the confusion in entity representations.
In the experiment, we used prompts such as ``[X] was founded in \_''. 
We compared the consistency of model responses by substituting mentions of two types of entities with different mention variability for [X].
As discussed in \S~\ref{subsec:Mention variability}, entities with low mention variability tend to have mentions encoded as a clustered representation, while entities with high mention variability show confusion between representations. 
The intuition behind this experimental setup is that differences in mention variability cause structural differences in representations, which affect variations in word prediction.
We sampled 20 organization entities each with low (0-0.3) and high (0.7-1.0) mention variability and predicted [Object] using the format ``[Subject] [Relation]'' with mentions belonging to these entities as Subject.\footnote{For Relations, we used ``[X] was founded in [Y].'' and ``The headquarter of [X] is in [Y].'' from LAMA Probe~\citep{petroni-etal-2019-language}. Llama-2 7B was used as the model.}
Word prediction consistency was calculated by collecting words for [Object] (the first word formed by combining generated tokens) and computing the proportion of the most frequent prediction within the same entity. 
Then, we calculated the average for both low and high mention variability groups.

The results revealed that while word prediction consistency averaged 71\% for low mention variability, it significantly decreased to an average of 39\% for high mention variability.
In other words, we confirmed that higher mention variability leads to lower model prediction consistency, supporting the hypothesis that the internal state of entity representations confusion causes prediction inconsistency.
A more detailed qualitative analysis of cases with inconsistent word prediction is presented in Appendix~\ref{subsec:appendix_output_inconsistency}.

\paragraph{Entity Disambiguation Accuracy} \label{subsubsec:entity_disambiguation}
We investigate how the degree of separation in entity representations affects entity disambiguation accuracy.
Intuitively, if entity representations are well-separated, the model should successfully perform entity disambiguation.
To examine the relationship between representation separation and entity disambiguation accuracy, we conducted in-context learning experiments as follows.
We input the following prompt to the LM: \texttt{Does \textbf{\textcolor{purple}{X}} refer to A or B? One-word answer only. A: San Diego B: San Francisco Answer:}.
We then use PatchScope~\citep{Patchscopes2024} to patch contextualized mention embeddings of San Diego or San Francisco into the position marked as \texttt{\textbf{\textcolor{purple}{X}}}.
If the entity representations of San Diego and San Francisco are well-separated, the model should correctly resolve the entity disambiguation task.
We used approximately 7,800 entities, consistent with the dataset size in \S~\ref{sec:result}.
We employed 10-shot learning and constrained the model to output only ``A'' or ``B'' as answers. We selected Llama-2 13B as it supports few-shot learning. Detailed experimental settings are provided in Appendix~\ref{subsec:appendix_entity_disambiguation}.

\begin{figure}[t]
\centering
\includegraphics[width=\linewidth]{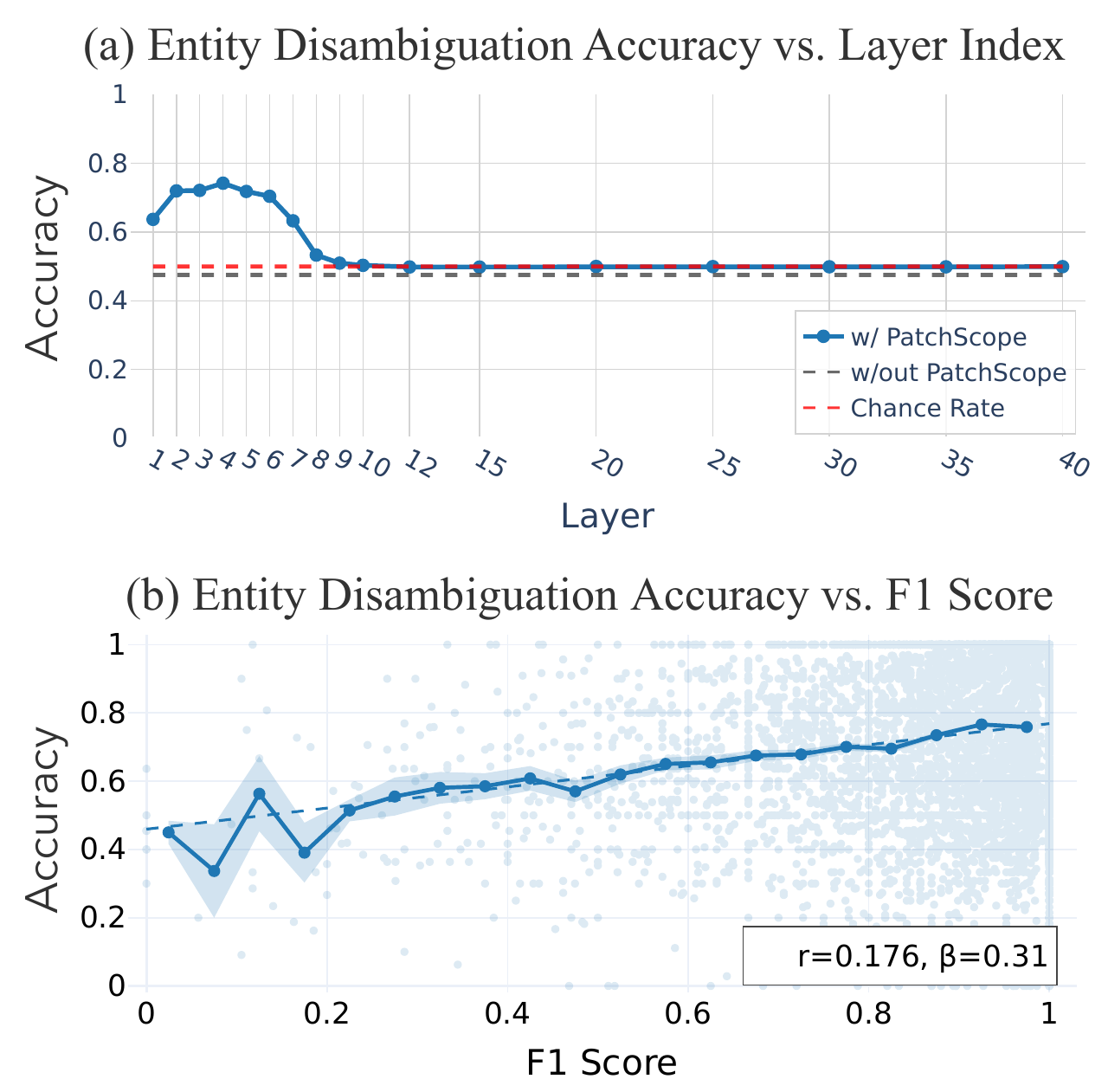}
\caption{
(a) Entity disambiguation accuracy for patch embeddings extracted from different Llama-2 13B layers.
(b) Entity disambiguation accuracy vs. representation separation (F1 score) for 4th layer patch embeddings from Llama-2 13B.
Entity disambiguation accuracies are averaged within each bin on the F1 score.
$r$: Pearson correlation, $\beta$: regression coefficient.
}
\label{fig:entity_disambiguation}
\end{figure}

Figure~\ref{fig:entity_disambiguation} presents our results. Figure~\ref{fig:entity_disambiguation}-(a) shows the effect of varying the layer used for patching. When using representations from early layers, entity disambiguation accuracy reaches approximately 74\%, significantly exceeding the 50\% chance rate. 
This finding Accuracy aligns with the trends observed in \S~\ref{subsec:representation_analysis}, suggesting that information from early layers contributes to entity disambiguation performance. 
Figure~\ref{fig:entity_disambiguation}-(b) plots entity disambiguation accuracy against F1 score using layer 4, which achieved the highest entity disambiguation accuracy. The Pearson correlation coefficient is approximately 0.18, with a regression coefficient $\beta$ of 0.31.
The data shows an upward trend, indicating that entities with well-separated representations tend to achieve higher entity disambiguation accuracy.

\section{Discussion: Insights from Isomorphism}
\label{sec:discussion}
In \S~\ref{sec:result}, we revealed the extent to which LMs can perform entity identification. 
Interpreting these experimental results from the perspective of isomorphism described in \S~\ref{subsec:Purity_IP_F1} and Table~\ref{tab:method_comparison} in the Appendix, we can conclude that LMs exhibit a higher degree of isomorphism between the structure of entity representations and the structure of entity classes compared to FastText and random baselines.
LMs achieved an AUC of approximately 0.9 in cases with mention ambiguity (\S~\ref{subsec:Mention ambiguity}) and an AUC of about 0.8 in cases with mention variability (\S~\ref{subsec:Mention variability}), both scores being significantly higher than baselines. 
Furthermore, we observed similarities in the structure of entity representations across different LMs (\S~\ref{subsubsec:rsa}).
These results experimentally support The Platonic Representation Hypothesis~\citep{The_Platonic_Representation_Hypothesis2024}.
Specifically, the similarity in entity representation structures across different LMs and their high isomorphism suggest that entity representation structures are ``converging'' toward a certain point, which can be interpreted as the structure of real-world entities (in this study, the discrete topology of ``mentions belonging to the same entity class'').
Our findings reinforce previous empirical observations that LMs can encode specific knowledge structures existing in the real-world through text-only learning~\citep{abdou-etal-2021-language, patel2022mapping, chen2023correlationlargelanguagemodels, gurnee2024language, DBLP:conf/iclr/HernandezSHMWAB24, park2024the}. 
Additionally, in \S~\ref{subsec:entity_representation_and_word_pred}, we demonstrated that the structure of representations affects both the consistency of word predictions and entity disambiguation performance. 
We consider this one of the positive aspects of achieving high isomorphism.

\section{Related Work}
\label{sec:related_work}
We review methods for verifying LMs knowledge and  position our research by organizing existing findings on knowledge structures.

\subsection{Methods for Analyzing LM Knowledge}
While there are numerous studies examining what knowledge LMs possess, their methodologies can be broadly categorized into three approaches.
\textbf{(i) Behavioral analysis}~\cite{petroni-etal-2019-language, jiang-etal-2020-know, marjieh2023large, patel2022mapping} explores knowledge encoded in LMs using prompts designed to elicit responses.
For example, evaluating whether the model can predict ``Hawaii'' from the prompt ``Barack Obama was born in \_''.
While this approach is interpretable to humans and provides intuitive insights, it has been reported that different phrasings of prompts with the same meaning can yield different prediction results~\cite{jiang-etal-2020-know, heinzerling-inui-2021-language}.
Additionally, treating the model as a black box makes it difficult to understand how specific input words are encoded within the LM.

\textbf{(ii) Classifier-based probing}~\cite{alain2017understanding, hewitt-manning-2019-structural, lietard-etal-2021-language, abdou-etal-2021-language, gurnee2024language} evaluates whether LMs' representations encode specific information by measuring the performance of trained classifiers.
These analysis results depend on classifier performance and can vary significantly based on optimizer and initialization settings~\cite{zhou-srikumar-2021-directprobe}.

\textbf{(iii) Representational analysis}~\cite{zhou-srikumar-2021-directprobe, abdou-etal-2021-language, park2024the,park2024iclrincontextlearningrepresentations} analyzes LMs' internal representations using unsupervised methods.
For instance, when measuring the similarity between two representational structures, it compares distance matrices between embedding pairs.
This approach can evaluate representations directly without classifiers, making it less susceptible to optimization effects and enabling a purer understanding of the model's internal structure.
The Purity~\cite{Zhao2001CriterionFF} and IP metrics used in this study belong to this representational analysis category.

\subsection{Structure in LM Representations}
The correspondence between LMs' representational spaces and real-world knowledge structures has been studied from various perspectives. \citet{mikolov-etal-2013-linguistic} demonstrated that word embeddings can linearly represent semantic relationships in vector space. Subsequently, \citet{petroni-etal-2019-language} and \citet{poerner-etal-2020-e} used behavioral analysis to prove that pre-trained LMs store factual knowledge composed of entities and relations. For instance, they confirmed that LMs could predict appropriate words like ``Hawaii'' from inputs such as ``Barack Obama was born in \_\_''. The observation that LMs could memorize knowledge through text-only learning and make such predictions led to various studies exploring LMs' internal structures, based on the assumption that their internal representations must encode some form of real-world knowledge structure. \citet{abdou-etal-2021-language} and \citet{patel2022mapping} showed that LMs' representations of color words resemble human color perception structures. \citet{chen2023correlationlargelanguagemodels} and \citet{gurnee2024language} demonstrated that geographical relationships between countries and cities are reflected in LMs' representational spaces. Furthermore, \citet{DBLP:conf/iclr/HernandezSHMWAB24} and \citet{park2024the} revealed that relationships between entities and hierarchical structures of concepts are preserved in the representational space. These studies suggest that LMs' representations capture real-world knowledge and relationships, providing important insights into what structures LMs can encode through text-only learning and where the limits of their representational capabilities lie~\citep{bender-koller-2020-climbing, bisk-etal-2020-experience, merrill-etal-2021-provable,DBLP:conf/iclr/MerulloCEP23, The_Platonic_Representation_Hypothesis2024}. Our research focuses on a fundamental question: whether LMs can distinguish mentions as entities. We analyzed this mention ambiguity and variability. Our findings further strengthen existing empirical evidence that LMs can acquire specific real-world knowledge structures through text-only learning.

\section{Conclusion}
We investigated how LMs encode entity mentions in their internal representations, focusing on the process of distinguishing entities in the presence of mention ambiguity and mention variability. Through experiments with five autoregressive LMs, these models achieved higher performance in entity identification compared to baselines. Our analysis revealed the degree of isomorphism between LM representations and real-world entity-centric knowledge structures, suggesting that LMs can effectively encode discrete entity relationships through text-only training. These findings deepen our understanding of how LMs internally represent and process entity information, providing new insights into the relationship between LM representation spaces and structural knowledge.

\newpage

\section*{Limitations}
There are primarily three limitations in this study. First, while the structure composed of entities is inherently language-independent (``Tokyo'', ``東京'', ``\Chinese{东京}'' refer to the same entity), our experiments only covered English. Therefore, it remains unclear whether the structure of entity representations is truly constructed in a language-independent manner. Important future work includes investigating whether the structure of entity representations is constructed in a language-independent way, and how such representational structures relate to practical applications (such as the success rate of cross-lingual transfer, where knowledge learned in one language is transferred to another).

Second, it is unclear what defines an adequate sample size for measuring the Purity and IP.
In measuring the Purity and IP, the criterion is based on whether the entity representations are mixed with ``other'' entity representations.
Consequently, the degree of mixing is relatively determined by the prepared data.
In other words, even if the LM genuinely confuses ``Barack Obama'' with ``Michelle Obama,'' it can be said that the confusion cannot be detected if either one is absent from the data.
In the experiments, we used about 160,000 sentences; however, while this is a substantial dataset, we cannot conclusively determine whether this sample size is sufficient to capture all potential entity confusions that might exist in the model.

Finally, perfect separation of representations may not be optimal or desirable across all scenarios. While our study assumes that entity-level representations should be distinct, there are scenarios where representation overlap could be beneficial. For example, when treating categories like painters, musicians, and novelists as single classes, representation overlap might be advantageous. Such overlap suggests that representations cluster around broader categories like ``artist,'' potentially encoding hierarchical categorical knowledge.
Furthermore, even in entity-level representation separation, it is debatable whether achieving perfect entity identification is a desirable property in all cases.
Taking entity identification to the extreme, a model that maps all mentions of an entity to exactly the same embedding would achieve perfect clustering scores.
In other contexts, the phenomenon where internal representations of instances within a class converge to a single point has been widely observed and is known as ``neural collapse'' \cite{hui2022limitationsneuralcollapseunderstanding}.
Research has shown that neural collapse during training isn't necessarily beneficial for all types of generalization and can sometimes be counterproductive. While it's unclear whether these findings directly apply to entity representations, if LMs use classification ($\approx$ entity identification) as an intermediate step for next-token prediction, we might expect to see neural collapse patterns emerge in intermediate layers. This raises the question of whether perfect entity identification is truly desirable. A critical direction for future work would be to investigate the potential drawbacks of having model representations that closely mirror real-world entity structures.

\section*{Ethics Statement}
In this study, we perform a novel analysis and investigate how LMs recognize knowledge about the world through internal representations.
In recent years, there has been a growing concern about the risk of socially harmful biases (e.g., racial or gender biases) in the text generated by LMs.
This study has the potential to contribute to a better understanding of this issue, by analyzing how LMs internally represent undesirable biases.

\section*{Acknowledgements}
We would like to thank the members of the Tohoku NLP Group for their insightful comments.
This work was supported by JSPS KAKENHI Grant Number JP21K17814, JP25KJ0628, JP22H05106, JP22H03654; JST CREST Grant Number JPMJCR20D2, Japan; JST ACT-X Grant Number JPMJAX200S, Japan; and JST FOREST Grant Number JPMJFR2331, Japan.

\bibliography{custom}

\clearpage
\appendix

\section{Details of the Purity and Inverse Purity Algorithm}
\label{sec:Algorithm}
The calculation of purity~\cite{Zhao2001CriterionFF} and inverse purity are as follows. Both metrics evaluate the quality of clusters with respect to their class distributions, with purity focusing on how well each cluster contains a single class, and inverse purity measuring how well the embeddings of each class are separated into distinct clusters.\footnote{
In Figure~\ref{fig:method}, the red-shaded area is referred to as the ``class,'' and the blue-shaded area is referred to as the ``cluster,'' which are used to calculate Equations~\ref{eq:local_purity} and \ref{eq:local_inverse_purity}.
We calculate the local purity and local IP by considering each embedding's ``class'' as the entity it refers to and each embedding's ``cluster'' as the nearest centroid.
In other words, the ``clusters'' refer to those formed by the Voronoi diagram using each centroid.
The F1 score, composed of Purity and IP, can be used to evaluate the degree to which embeddings of entities belonging to the same class are grouped into a single cluster.}

\begin{table*}[t]
\small
\centering
\begin{tabular}{lll}
\toprule
\begin{tabular}[c]{@{}l@{}}
Structure \\            
\end{tabular} 
& 
\begin{tabular}[c]{@{}l@{}}
Requirement: Desired correspondence between\\
\ColoredText{orange!20}{real-world structure} and \ColoredText{teal!15}{representation space structure}
\end{tabular}
& \begin{tabular}[c]{@{}l@{}}Soft Evaluation \end{tabular} \\ \midrule
\begin{tabular}[c]{@{}l@{}} 
Metric Space \\
\citep{abdou-etal-2021-language}
\end{tabular}
& 
\begin{tabular}[c]{@{}l@{}}
$\forall x_1, x_2, x_3, x_4 \in X$ \\
\ColoredText{orange!20}{$d_X(x_1, x_2) < d_X(x_3, x_4)$} \\
$\Leftrightarrow$
\ColoredText{teal!15}{$d_X(h_1, h_2) < d_X(h_3, h_4)$} 
\\
\\
$\forall x_1, x_2, x_3, x_4 \in X$ \\
\ColoredText{orange!20}{$\mathbbm{1}[d_X(x_1, x_2) < d_X(x_3, x_4)]$} \\
$=$ \ColoredText{teal!15}{$\mathbbm{1}[d_X(h_1, h_2) < d_X(h_3, h_4)]$}
\\
\\
\includegraphics[width=0.4\textwidth]{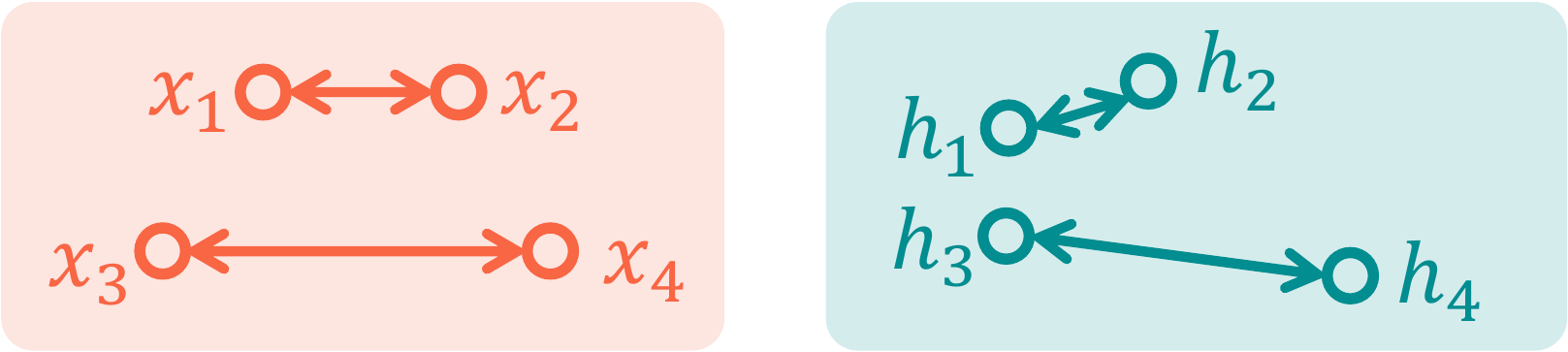}
\\
Whether distance relationships are preserved
\end{tabular} 
& 
\begin{tabular}[c]{@{}l@{}}
$\mathbb{P}_{x_1, x_2, x_3, x_4 \sim X}[$ \\
$\mathbbm{1}[$
\ColoredText{orange!20}{$d_X(x_1, x_2) < d_X(x_3, x_4)$}$]$ \\
$= \mathbbm{1}[$
\ColoredText{teal!15}{$d_X(h_1, h_2) < d_X(h_3, h_4)$}$]$ \\
$]$
\\
\\
This roughly corresponds to \\Kendall's $\tau$, 
which in turn \\approximately corresponds \\
to RSA~\citep{RSA}.
\end{tabular} 
\\ \midrule
\begin{tabular}[c]{@{}l@{}}
Discrete Topology \\
(Ours)   
\end{tabular}
& 
\begin{tabular}[c]{@{}l@{}}
$\forall x_1, x_2$ \\
\ColoredText{orange!20}{$x_1 = x_2$} \\
$\Leftrightarrow$ 
\ColoredText{teal!15}{$h_1 = h_2$}
\\
\\
$\forall x_1, x_2$ \\
\ColoredText{orange!20}{$\mathbbm{1}[x_1 = x_2]$} \\
$=$
\ColoredText{teal!15}{$\mathbbm{1}[h_1 = h_2]$}
\\
\\
\includegraphics[width=0.4\textwidth]{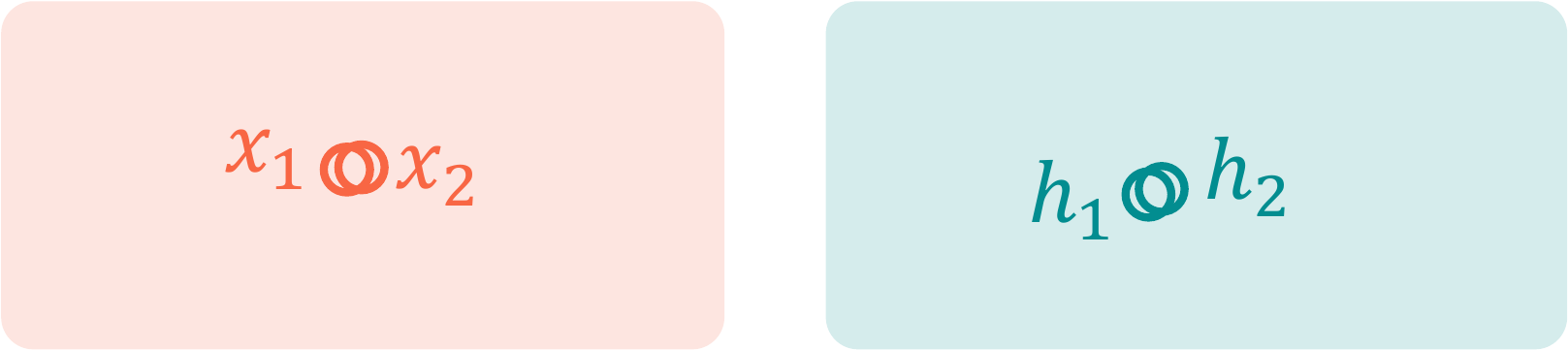}
\\
Whether the same entity occupies the same position
\end{tabular}
& 
\begin{tabular}[c]{@{}l@{}}
$\mathbb{P}_{x_1,x_2 \sim X}[$ \\
$\mathbbm{1}[$
\ColoredText{orange!20}{$x_1 = x_2$}$]$ \\
$= \mathbbm{1}[$
\ColoredText{teal!15}{$\mathrm{Cluster}(h_1) = \mathrm{Cluster}(h_2)$}$]$
$]$
\\
\\
This approximately corresponds\\ to 
the F1 score composed of Purity\\ and IP.
By introducing clustering,\\ we relax the
requirement of exact \\position matching
to proximity-based\\ matching.
\end{tabular} 
\\ \bottomrule
\end{tabular}
\caption{Comparison between existing Isomorphic analysis and our approach.}
\label{tab:method_comparison}
\end{table*}

\paragraph{Local Purity}

We begin by calculating the purity for each cluster, referred to as the local purity. For each cluster $\mathrm{Cluster}(e)$, we compute its $\mathrm{local\;Purity}(e)$ by following these steps:

\begin{enumerate}

\item For each entity $e \in \mathcal{E}$ (e.g., ''Biden''), compute the centroid $\bm{b}_e \in \mathbb{B}$ of its mention embeddings $\bm{X}_e = \{\bm{x}_e^{1}, \bm{x}_e^{2}, \dots\}$:

    \begin{align}
        \bm{b}_e \coloneqq \frac{1}{\lvert \bm{X}_e \rvert} \sum_{\bm{x}_e \in \bm{X}_e} \bm{x}_e
        \text{.}
    \end{align}

    \item For each embedding $\bm{x}$, determine the $\mathrm{Cluster}(e)$ it belongs to by assigning it to the nearest centroid $\bm{b}_e$:

    \begin{align}
        \mathrm{Cluster}(e) \coloneqq \{\bm{x} \mid \bm{b}_e = \argmin_{\bm{b}' \in \mathbb{B}} d(\bm{x}, \bm{b}')\}
        \label{eq:cluster}
        \text{,}
    \end{align}

    where \( d(\bm{x}, \bm{y}) \) represents the distance function.
    In our experiments, we employed Euclidean distances, which are commonly used in NLP.
    This step is equivalent to an intermediate stage of k-means clustering.

    \item Define the gold class for entity $e$ as:

    \begin{align}
        \mathrm{Class}(e) \coloneqq \bm{X}_e
        \label{eq:class}
        \text{.}
    \end{align}

    \item Compute the local purity of the $\mathrm{Cluster}(e)$ by calculating the fraction of embeddings in the cluster that belong to the most frequent class $\hat{e}$:

    \begin{align}
        \mathrm{local\;Purity}(e) \coloneqq \frac{\lvert \mathrm{Cluster}(e) \cap \mathrm{Class}(\hat{e}) \rvert}{\lvert \mathrm{Cluster}(e) \rvert}
        \label{eq:local_purity}
        \text{,}
    \end{align}

    where $\mathrm{Class}(\hat{e})$ is the most frequent class in the cluster, and $\lvert \mathrm{Cluster}(e) \rvert$ is the total number of embeddings in the $\mathrm{Cluster}(e)$.
\end{enumerate}

\paragraph{Purity}

The overall purity, which measures how well the clusters contain a single class, is calculated as the weighted average of the local purities:

\begin{align}
    \mathrm{Purity}(\mathcal{E}) \coloneqq \frac{1}{N} \sum_{e \in \mathcal{E}} \mathrm{local\;Purity}(e) \lvert \mathrm{Cluster}(e) \rvert
    \label{eq:purity}
    \text{,}
\end{align}

where $N$ is the total number of mention embeddings across all clusters.

\paragraph{Local Inverse Purity (Local IP)}

Next, we calculate the inverse purity for each entity, referred to as the local inverse purity. For each entity $e$, we compute $\mathrm{local\;IP}(e)$ by following these steps:

\begin{enumerate}
    \item We use the definitions of $\mathrm{Cluster}(e)$ from Equation~\ref{eq:cluster} and $\mathrm{Class}(e)$ from Equation~\ref{eq:class} for each entity $e$.
    \item Compute the local inverse purity $\mathrm{local\;IP}(e)$ by calculating the fraction of embeddings from the gold class $\mathrm{Class}(e)$ that are contained within the cluster $\mathrm{Cluster}(e)$:

    \begin{align}
        \mathrm{local\;IP}(e) \coloneqq \frac{\lvert \mathrm{Cluster}(e) \cap \mathrm{Class}(e) \rvert}{\lvert \mathrm{Class}(e) \rvert}
        \label{eq:local_inverse_purity}
        \text{,}
    \end{align}

    where $\lvert \mathrm{Class}(e) \rvert$ is the total number of embeddings in the gold class $\mathrm{Class}(e)$.
\end{enumerate}

\paragraph{Inverse Purity (IP)}

The overall inverse purity, representing the average degree of separation across all classes, is calculated as:

\begin{align}
    \mathrm{IP}(\mathcal{E}) \coloneqq \frac{1}{N} \sum_{e \in \mathcal{E}} \mathrm{local\;IP}(e) \lvert \mathrm{Class}(e) \rvert
    \label{eq:inverse_purity}
    \text{,}
\end{align}

where $N$ is the total number of mention embeddings.

This method provides a simple yet effective way to measure the embedding space in terms of both class containment (purity) and class separation (inverse purity).

\paragraph{F1 Score with Purity and Inverse Purity}
To further evaluate the embedding space, we calculate an F1 score using both purity and inverse purity. 
The F1 score provides a balanced measure that considers both the ability of the clusters to contain single classes (purity) and the degree of separation of classes into distinct clusters (inverse purity). The F1 score is computed as the harmonic mean of purity and inverse purity, formulated as follows:

\begin{align} \mathrm{F1\;Score}(\mathcal{E}) \coloneqq 2 \cdot \frac{\mathrm{Purity}(\mathcal{E}) \cdot \mathrm{IP}(\mathcal{E})}{\mathrm{Purity}(\mathcal{E}) + \mathrm{IP}(\mathcal{E})} \label{eq:f1} \text{.} \end{align}

Here, the F1 score effectively balances the trade-off between purity and inverse purity. 
A high F1 score indicates that the clusters not only contain embeddings from the same class but also that embeddings from each class are well-separated into different clusters.

\section{Details of the Experimental Setup}
\label{sec:appendix}

\begin{table*}[th]
\centering
\small
\begin{tabular}{lll}
\toprule
DIFFICULTY & Low-difficulty examples & High-difficulty examples \\ 
\cmidrule(lr){1-1} \cmidrule(lr){2-2} \cmidrule(lr){3-3}
\begin{tabular}[c]{@{}l@{}}
\textbf{Mention ambiguity}: \\
Entropy of mention to \\ entity candidate mapping.
\end{tabular} & 
\begin{tabular}[c]{@{}l@{}}
\textbf{Mention}: `Ohio'\\ \\ 
\textbf{Entity candidate}:\\ 
\{Ohio: 198, Ohio River: 12\}
\end{tabular} & 
\begin{tabular}[c]{@{}l@{}}
\textbf{Mention}: `Oxford'\\ \\ 
\textbf{Entity candidate}:\\ 
\{Oxford: 26, University of Oxford: 13,\\ 
Oxford, Mississippi: 6, \\ 
Oxford (UK Parliament constituency): 6\}
\end{tabular} \\
\cmidrule(lr){1-1} \cmidrule(lr){2-2} \cmidrule(lr){3-3}
\begin{tabular}[c]{@{}l@{}}
\textbf{Mention variability}: \\
Average surface form \\ dissimilarity of pairs of \\
mention candidate \\owned by an entity.
\end{tabular} & 
\begin{tabular}[c]{@{}l@{}}
\textbf{Entity}: Emmy Awards\\ \\ 
\textbf{Mention candidate}:\\ 
(`Emmy', `Emmy Awards',\\ 
`Emmy award', `Emmy awards' \\ 
`Emmys')
\end{tabular} & 
\begin{tabular}[c]{@{}l@{}}
\textbf{Entity}: Māori All Blacks\\ \\ 
\textbf{Mention candidate}:\\ 
(`New Zealand Maori Rugby', \\
`Māori All Blacks', `NZ Maori', \\
`New Zealand Māori rugby union team', ...)
\end{tabular} \\ 
\bottomrule
\end{tabular}
\caption{Difficulty factors considered to potentially influence entity identification, with definitions and examples.
Numbers in the entity candidate lists represent the frequency of occurrence for each entity.
Detailed definitions of these factors are provided in Appendix~\ref{subsec:appendix_definition_mention_ambiguity},~\ref{subsec:appendix_definition_mention_variability}.
}
\label{tab:factors}
\end{table*}

\subsection{Pre-trained Language Models}
\label{sec:appendix_model}
In our experiments, we used the models shown in Table~\ref{tab:models_detail}.
We used the model provided in \url{https://github.com/huggingface/transformers}.

\begin{table}[htbp]
\centering
\small
\begin{tabular}{lccc}
\toprule
Models                  & Hidden dim. & \#Layer & \#Head \\ 
\cmidrule(lr){1-1}  \cmidrule(lr){2-2} \cmidrule(lr){3-3} \cmidrule(lr){4-4}
%BERT-tiny              & 128         & 2      & 2     \\
%BERT-mini              & 256         & 4      & 4     \\
%BERT-small             & 512         & 4      & 8     \\
%BERT-medium            & 512         & 8      & 8     \\
%BERT-base              & 768         & 12     & 12     \\
%BERT-large             & 1024        & 24     & 16     \\
%RoBERTa-base            & 768         & 12      & 12     \\
%RoBERTa-large           & 1024        & 24      & 16     \\
%ALBERT-base-v2          & 768         & 12      & 12     \\
%ALBERT-large-v2         & 1024        & 24      & 16     \\
%ALBERT-xxlarge-v2       & 4096        & 12      & 64     \\
%DistilBERT-base          & 768         & 6       & 12     \\ 
%LUKE-base               & 768         & 12      & 12     \\
%LUKE-large              & 1024        & 24      & 16     \\ 
GPT-2           & 768  & 12 & 12 \\
Llama-2-7B      & 4096 & 32 & 32 \\
Llama-2-13B     & 5120 & 40 & 40 \\
% Llama-2-70B     & 8192 & 80 & 64 \\
Llama-3-8B      & 4096 & 32 & 32 \\
% Vicuna-7B-v1.5  & 4096 & 32 & 32 \\
%Vicuna-13B-v1.5	& 5120 & 40 & 40 \\
%Pythia-6.9B	    & 4096 & 32 & 32 \\
%Pythia-12B      & 5120 & 36 & 40 \\
Mistral-7B-v0.3      & 4096 & 32 & 32 \\
% Falcon      & - & - & - \\

\bottomrule
\end{tabular}
\caption{Hyperparameters of each model's architecture.
}
\label{tab:models_detail}
\end{table}

\subsection{Probe Training Hyperparameters}
\label{subsec:appendix_probe_setting}
In \S~\ref{subsec:representation_analysis}, we performed linear probing using a single linear layer. 
The hyperparameters used for the probe training are shown in Table~\ref{tab:probe_Hyperparameters}. We split the data into training and test sets with a ratio of 8:2 using stratified sampling, and performed 3-fold cross-validation with early stopping monitoring the training loss.

\begin{table}[htbp]
\centering
\small
\begin{tabular}{lccc}

\toprule
 Optimizer          &  Adam~\citep{kingma2015adam}  \\
 Learning rate      & $1.0\times10^{-3}$ (constant) \\
 Batch size         & 1,024 \\
 Epochs             &  1,000 \\
\bottomrule

\end{tabular}
\caption{Hyperparameters of the trained probe.}
\label{tab:probe_Hyperparameters}
\end{table}

\subsection{Statistical Information of Experimental Data}
\label{subsec:appendix_data}
The statistical information of the data used in the experiments is shown in Table~\ref{tab:data}.
\begin{table*}[t]
\centering
\small
\begin{tabular}{lrrr}
\toprule
 ZELDA-TRAIN Corpus                                     &  \# Entities &  \# Mentions &  \# Sentences \\ 
 \cmidrule(lr){1-1}  \cmidrule(lr){2-2} \cmidrule(lr){3-3} \cmidrule(lr){4-4}
 All data                                                 &  7,877       &  24,289      &  164,516     \\
Ambiguous mention Subset &  1,386       &  602      &  30,082       \\
 %Entity-mention: one-to-one, Entity-context: one-to-many  &  7,887       &  7,887       &  164,590      \\ 
 \bottomrule
\end{tabular}
\caption{Statistics of the experiment dataset.}
\label{tab:data}
\end{table*}

\subsection{Detailed Definition of Mention Ambiguity}
\label{subsec:appendix_definition_mention_ambiguity}
Mention ambiguity is calculated as the entropy of the frequency distribution of entity candidates for a given mention.
Specifically, let \( N \) be the number of entity candidates corresponding to a mention, and \( F_i \) be the frequency of occurrence for each candidate \( i \). The mention ambiguity is then calculated through the following steps.

First, the total frequency \( F_{\text{total}} \) of all entity candidates for a given mention is expressed as:
\begin{align}
F_{\text{total}} = \sum_{i=1}^{N} F_i
\end{align}
Next, the relative frequency \( p_i \) for each entity candidate \( i \) is defined as the ratio of the candidate's frequency \( F_i \) to the total frequency \( F_{\text{total}} \):
\begin{align}
p_i = \frac{F_i}{F_{\text{total}}}, \quad i = 1, 2, \dots, N
\end{align}
Finally, mention ambiguity is represented by the entropy \( H \), defined by the following equation:
\begin{align}
H = - \sum_{i=1}^{N} p_i \log p_i
\end{align}
Using this entropy measure allows for quantitative evaluation of the degree of ambiguity associated with a given mention. When mention ambiguity is high, the difficulty of identifying the appropriate entity increases.
Conversely, when mention ambiguity is low, it indicates that the corresponding entity candidates are clear and there is little ambiguity.

\subsection{Detailed Definition of Mention Variability}
\label{subsec:appendix_definition_mention_variability}
Mention variability is a metric that quantifies the surface-level diversity of multiple mentions referring to the same entity.
In this research, we formulate this diversity as a dissimilarity measure based on Levenshtein distance.
Given a set of mentions $M_e = {m_1, m_2, ..., m_{|M_e|}}$ for entity $e$, the surface form dissimilarity $D$ is defined as follows:
\begin{align}
D &= 1 - \frac{2}{|M_e|(|M_e|-1)} \nonumber \\
&\quad\sum_{i=1}^{|M_e|-1} \sum_{j=i+1}^{|M_e|} 
\left(1 - \frac{L(m_i, m_j)}{\max(|m_i|, |m_j|)}\right)
\end{align}

\begin{align}
= \frac{2}{|M_e|(|M_e|-1)} \sum_{i=1}^{|M_e|-1} \sum_{j=i+1}^{|M_e|} \frac{L(m_i, m_j)}{\max(|m_i|, |m_j|)}
\end{align}
Here, $L(m_i, m_j)$ represents the Levenshtein distance between mentions $m_i$ and $m_j$, and $|m_i|$ denotes the length of mention $m_i$.
The derivation of the Levenshtein distance is described in \S~\ref{sec:appendix_lev}.
Using this surface form dissimilarity $D$, we can quantitatively evaluate the surface-level diversity (mention variability) of mentions associated with a given entity.
A higher value of $D$ indicates greater surface-level diversity of mentions.
Conversely, a lower value of $D$ suggests lower mention diversity and more similar surface forms.

\subsection{Derivation of the Levenshtein Distance}
\label{sec:appendix_lev}
The Levenshtein distance~\cite{levenshtein-dist} measures the dissimilarity between two strings, i.e., the higher the value, the less similar the pair.
This distance is calculated as the minimum number of single-character edits  (comprising insertions, deletions, or substitutions) necessary to change one string into the other. %
Given two strings $a$, $b$ (of length $|a|$ and $|b|$ respectively), the Levenshtein distance $L(a, b)$ is derived as follows:
\begin{align}
    L(a, b) = 
    1+\min \begin{cases}
    L(\text{tail}(a), b) \\
    L(a, \text{tail}(b)) \\
    L(\text{tail}(a), \text{tail}(b)) 
    \end{cases} 
\end{align}
Where the $\text{tail}(s)$ is the strings composed of the string following the except for the first character of the strings $s$.
Note that when $|a|=0$, $L(a, b)$ returns $|b|$, and when $|b|=0$, $L(a, b)$ returns $|a|$.
Additionally, if $s[n]$ is the $n$th character of the string $s$, and $a[0]=b[0]$, the $L(a, b)$ returns $L(\text{tail}(a), \text{tail}(b))$.

\subsection{Entity Representation Structural Similarity}
\label{subsec:appendix_rsa}

We conduct Representational Similarity Analysis (RSA) to examine the structural similarity of entity representations across different language models and compare them against baseline embeddings. This analysis reveals how different models organize semantic knowledge about entities in their representational space.

\subsubsection{Embedding Extraction Setup}

\textbf{Model Selection:} We analyze entity representations from seven different embedding sources: Llama-2-7B (layer 8), Llama-2-13B (layer 9), Llama-3-8B (layer 8), Mistral-7B (layer 8), and GPT-2 (layer 6). We use FastText wiki-news-300d-1M-subword and random embeddings as baseline comparisons.

\textbf{Layer Selection:} For transformer-based models, we extract representations from the layers that achieved the highest F1 scores in our main experiments (\S\ref{sec:result}). 

\textbf{Dimensionality Reduction:} All embeddings are projected to a 20-dimensional space using Linear Discriminant Analysis (LDA) to enable fair comparison across models with different native dimensionalities.

\textbf{Data Preprocessing:} We use the Zelda dataset. Due to computational memory constraints, we randomly sample 100,000 instances from the dataset for the RSA experiments.
\subsubsection{RSA Computation Methodology}

\textbf{Representational Similarity Matrix (RSM) Construction:} For each model's entity embeddings $\mathbf{E} \in \mathbb{R}^{n \times d}$, where $n$ is the number of entity mentions and $d=20$ is the embedding dimension, we compute pairwise similarities based on Euclidean distances:

\begin{equation}
\text{RSM}_{ij} = 1 - \frac{||\mathbf{e}_i - \mathbf{e}_j||_2}{\max_{k,l} ||\mathbf{e}_k - \mathbf{e}_l||_2}
\end{equation}

\textbf{RSA Score Calculation:} We compute the representational similarity between models by correlating the upper triangular portions of their RSMs (excluding diagonal elements). We use Spearman rank correlation to capture monotonic relationships while being robust to outliers:

\begin{align}
\text{RSA}(M_1, M_2) &= \rho_{\text{Spearman}}(\text{flatten}(\text{triu}(\text{RSM}_1)), \nonumber \\
&\quad \text{flatten}(\text{triu}(\text{RSM}_2)))
\end{align}

where $\text{triu}()$ extracts the upper triangular matrix and $\text{flatten}()$ converts it to a vector.

\subsection{Entity Disambiguation with Patchscopes}
\label{subsec:appendix_entity_disambiguation}
\paragraph{Model Configuration}
We conduct our entity disambiguation experiments using Llama-2-13B.
The same model is used both for extracting hidden states and for inference with patched representations.

\paragraph{Data Preparation}
Our experimental data consists of:
\begin{itemize}
    \item \textbf{Input data}: Following the main experiments in \S~\ref{sec:result}, we use approximately 7,800 entities and 160,000 mention embeddings.
    \item \textbf{Embedding configuration}: 20-dimensional LDA embeddings.
\end{itemize}

\paragraph{Patchscopes Configuration}
We employ the Patchscopes~\citep{Patchscopes2024} with the following specifications:
\begin{itemize}
    \item \textbf{Source representation}: Final token representation from the entity mention in context, consistent with our main experiments in \S~\ref{sec:result}.
    \item \textbf{Target prompt}: Binary choice format: 10-shot examples + ``Does X refer to A or B? One-word answer only. A: \{option\_a\} B: \{option\_b\} Answer:''
    \item \textbf{Intervention layers}: We analyze all layers to understand how entity representations evolve across model depth. When extracting hidden states from layer $\ell$ of the source context, we patch them into the corresponding layer $\ell^{*}$ of the target prompt (where $\ell = \ell^{*}$). 
    \item \textbf{Position indices}: 
    \begin{itemize}
        \item Source position: -1 (last token of the mention)
        \item Target position: Automatically detected X token in the prompt template
    \end{itemize}
\end{itemize}

\paragraph{Prompt Design}
We use a standardized prompt template with 10-shot in-context learning examples. This prompt design aims to constrain the model's output to either ``A'' or ``B''. By showing the model 10 examples, we successfully guide it to output only ``A'' or ``B'' as answers.

\begin{verbatim}
Does Michael Jordan refer to A or B? 
One-word answer only.
A: Spain
B: Michael Jordan
Answer: B
[... 9 more examples ...]
Does X refer to A or B? 
One-word answer only.
A: {option_a}
B: {option_b}
Answer:
\end{verbatim}

The assignment of correct entity vs.\ distractor to options A/B is deterministically randomized using MD5 hashing of the mention-entity pair to ensure balanced label distribution.

\paragraph{Distractor Selection}
We select distractors as the nearest neighbor entities to X's embedding (contextualized mention embeddings from Llama-2-13B). An actual example is shown below:

\begin{verbatim}
Does X refer to A or B? 
One-word answer only.
A: San Diego
B: San Francisco
Answer:
\end{verbatim}
This approach is based on the intuition that if the representations are mixed with those of neighboring entities, the entity disambiguation task will likely fail.
Using nearest neighbors as distractors provides categorically similar alternatives, creating an appropriately challenging task compared to random distractor selection. For instance, the nearest neighbor to ``Tesla, Inc.'' is ``Nissan'', both being automobile manufacturers, making the discrimination task non-trivial yet meaningful.

\paragraph{Baselines}
The baselines are chance rate and w/out PatchScope.
Since this task is binary choice, the chance rate is 50\%.
w/out PatchScope does not patch entity representations to ``\texttt{\textbf{\textcolor{purple}{X}}}''.
That is, it inputs the example and \texttt{Does \textbf{\textcolor{purple}{X}} refer to A or B? One-word answer only. A: San Diego B: San Francisco Answer:} for in-context learning (ICL).
w/out PatchScope is expected to perform nearly at chance level, and indeed, Figure~\ref{fig:entity_disambiguation}~(a) confirms that it performs equivalently to the chance rate.

\section{Details of the Analysis Results}
\label{sec:appendix_detail_analysis_results}

\subsection{Dimensions and F1 Score}
\label{sec:appendix_dimensions_f1_score}
\begin{table*}[htbp]
\centering
\small
\begin{tabular}{@{}rrrrrrrr@{}}
\toprule
Dimension & Llama-2-7B & Llama-2-13B & Llama-3-8B & Mistral-7B & GPT-2 & FastText & \begin{tabular}[t]{@{}l@{}}Random\\Embeddings\end{tabular} \\
\midrule
1         & 0.01       & 0.01        & 0.00       & 0.01       & 0.00  & 0.23     & 0.00              \\
2         & 0.07       & 0.07        & 0.06       & 0.06       & 0.05  & 0.38     & 0.00              \\
5         & 0.51       & 0.51        & 0.52       & 0.49       & 0.46  & 0.68     & 0.00              \\
10        & 0.80       & 0.80        & 0.81       & 0.78       & 0.77  & 0.80     & 0.00              \\
20        & 0.91       & 0.91        & 0.91       & 0.90       & 0.85  & 0.85     & 0.01              \\
30        & 0.93       & 0.93        & 0.93       & 0.92       & 0.87  & 0.87     & 0.01              \\
40        & 0.94       & 0.94        & 0.94       & 0.93       & 0.87  & 0.87     & 0.02              \\
50        & 0.94       & 0.94        & 0.94       & 0.94       & 0.88  & 0.88     & 0.05              \\
60        & 0.95       & 0.95        & 0.95       & 0.94       & 0.88  & 0.88     & 0.05              \\
80        & 0.95       & 0.95        & 0.95       & 0.95       & 0.88  & 0.88     & 0.08              \\
100       & 0.95       & 0.95        & 0.95       & 0.95       & 0.88  & 0.88     & 0.10              \\
200       & 0.96       & 0.96        & 0.95       & 0.96       & 0.88  & 0.88     & 0.30              \\
300       & 0.96       & 0.96        & 0.95       & 0.96       & 0.88  & 0.85     & 0.48              \\
500       & 0.96       & 0.96        & 0.96       & 0.96       & 0.90  & --       & 0.69              \\
1000      & 0.96       & 0.96        & 0.96       & 0.96       & --    & --       & 0.87              \\
2000      & 0.96       & 0.96        & 0.96       & 0.96       & --    & --       & 0.96              \\
3000      & 0.96       & 0.96        & 0.96       & 0.96       & --    & --       & 0.98              \\
4096      & 0.93       & 0.93        & 0.93       & 0.94       & --    & --       & 0.99              \\
\bottomrule
\end{tabular}
\caption{F1 scores corresponding to the number of dimensions for each model.}
\label{tab:f1_dimensions}
\end{table*}
The correspondence between the number of dimensions and F1 scores for each model is shown in Table~\ref{tab:f1_dimensions}.

\subsection{Mention Ambiguity}
\label{sec:appendix_mention_ambiguity}

\begin{table}[ht]
\centering
\small
\begin{tabular}{@{\extracolsep{\fill}}lcc@{}}
\toprule
\textbf{Method} & \begin{tabular}[c]{@{}c@{}}\textbf{Mention}\\ \textbf{Ambiguity}\end{tabular} & \begin{tabular}[c]{@{}c@{}}\textbf{Mention}\\ \textbf{Variability}\end{tabular} \\
\midrule
Random Embeddings & 0.666 & 0.003 \\ \cmidrule(lr){1-1}  \cmidrule(lr){2-2} \cmidrule(lr){3-3} 
\begin{tabular}[c]{@{}l@{}}Unique Mention\\ Embeddings\end{tabular} & 0.375 & 0.413 \\  \cmidrule(lr){1-1}  \cmidrule(lr){2-2} \cmidrule(lr){3-3} 
FastText & 0.352 & 0.611 \\
\bottomrule
\end{tabular}
\caption{AUC scores for baseline embedding methods. 
These static embeddings serve as comparison points for the contextualized representations from transformer models shown in Tables~\ref{tab:auc_mention_amb} and~\ref{tab:auc_mention_vari}.
All embeddings are reduced to 20 dimensions. 
}
\label{tab:baseline_auc}
\end{table}

\subsubsection{Model Comparison Results}
\label{sec:appendix_mention_ambiguity_models} 
\begin{figure*}[t]
\centering
\includegraphics[width=16cm]{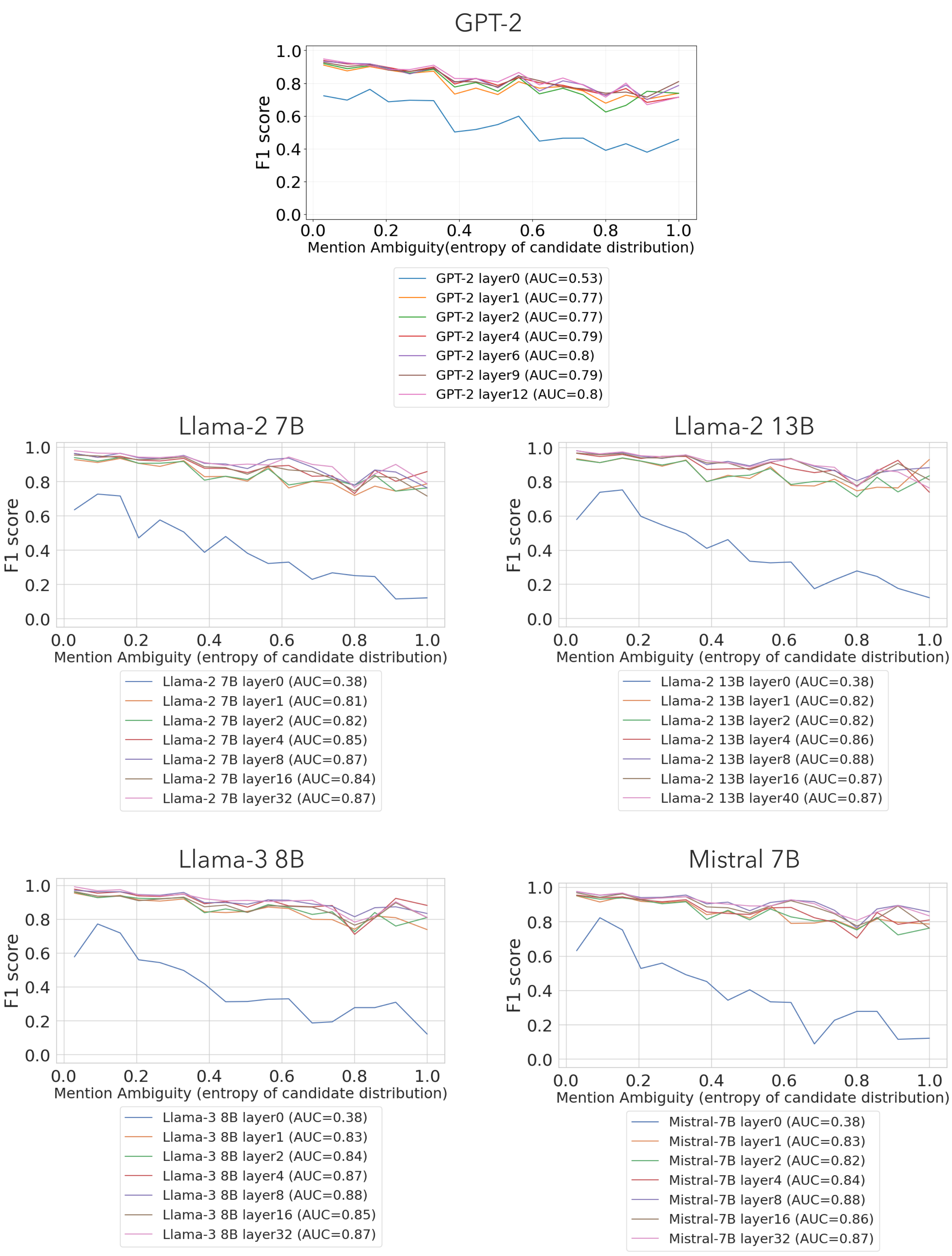}
\caption{
Results for all models: mention ambiguity and the separability of entity representations (F1 score). 
}
\label{fig:mention_ambiguous_result_summary_all_model}
\end{figure*}

Table~\ref{tab:baseline_auc} shows the AUC of mention ambiguity and F1 score for the baseline. Table~\ref{tab:auc_mention_amb} and Figure \ref{fig:mention_ambiguous_result_summary_all_model} show the analysis results from layer 0 to the final layer for mention ambiguity using GPT-2, Llama-2 7B, Llama-2 13B, Llama-3 8B, and Mistral 7B.

\subsubsection{Results with Average Subword Token Embeddings}
\label{subsec:appendix_mention_amb_ave}

\begin{figure*}[t]
\centering
\includegraphics[width=\linewidth]{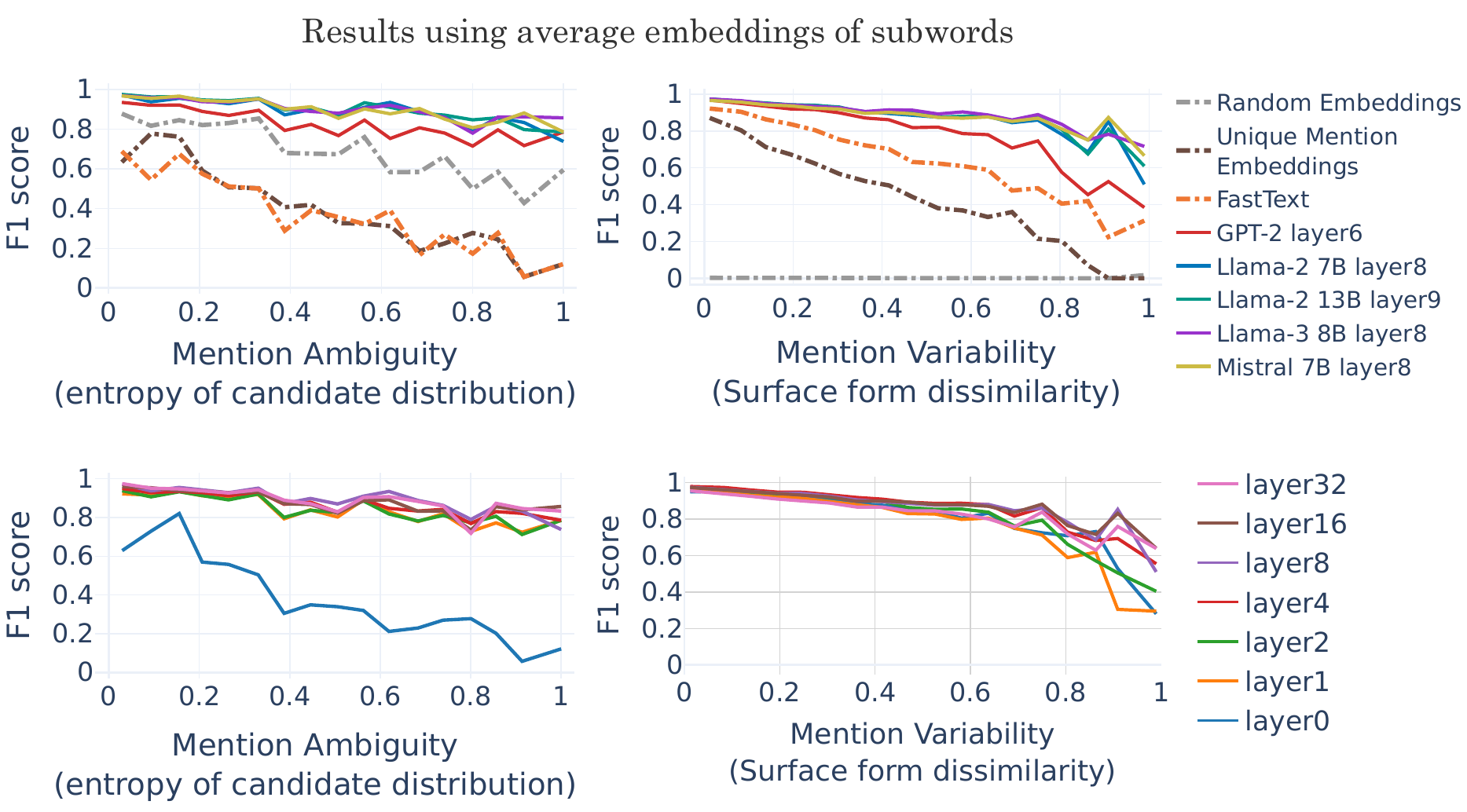}
\caption{
Separability (F1 scores) of entity representations using average embeddings of subword tokens for each difficulty~(left: mention ambiguity, right: mention variability). 
F1 scores are averaged within each bin on the x-axis. 
Top: Results across all models, using the layer with highest AUC for LM scores. 
Bottom: Layer-wise analysis for Llama-2 7B. 
}
\label{fig:result_summary_ave}
\end{figure*}

Figure~\ref{fig:result_summary_ave} shows the results when using average embeddings of subword tokens as the target tokens for analysis.
We can observe that the trends are nearly identical to those in Figure~\ref{fig:result_summary} in \S~\ref{sec:result}, which used last token embeddings.

\begin{figure*}[t]
\centering
\includegraphics[width=16cm]{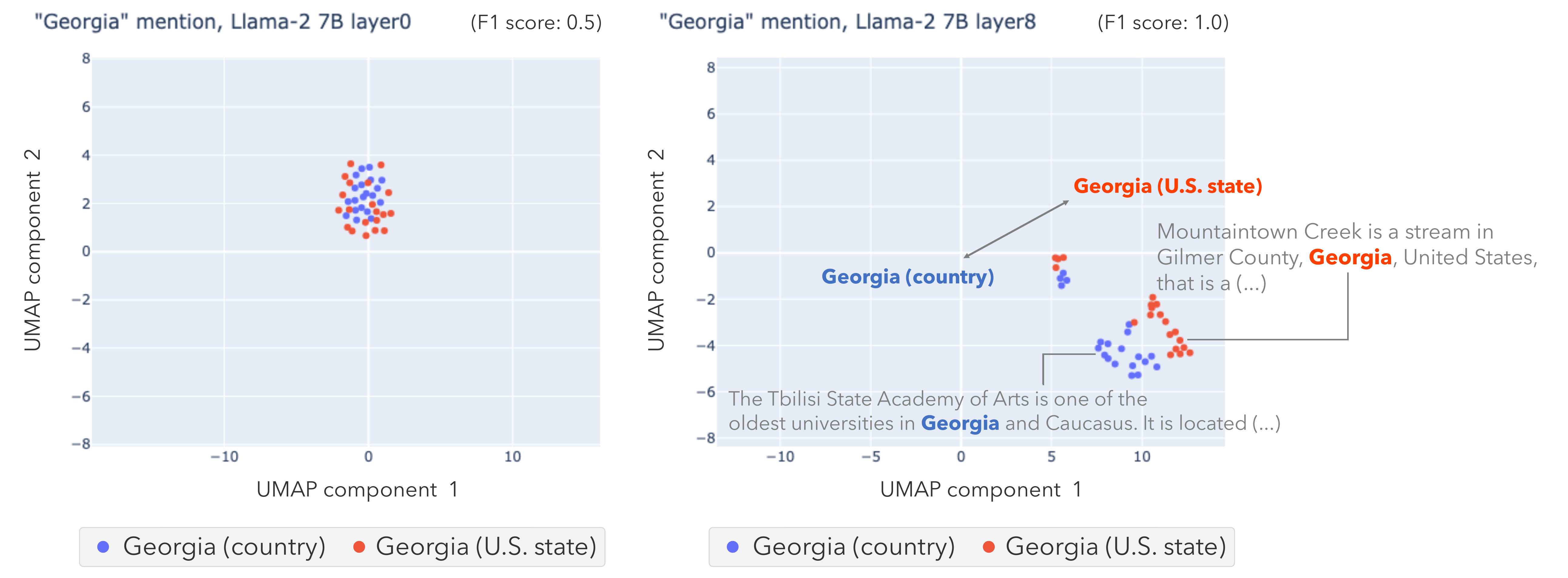}
\caption{
UMAP visualization of the ambiguous mention ``Georgia.'' In layer 0 of Llama-2 7B, the embeddings are mixed, but in layer 8, the embeddings are grouped by entity.
}
\label{fig:mention_ambiguous_qualitative_layer-wise_Georgia}
\end{figure*}

\subsection{Mention Variability}
\label{sec:appendix_mention_variability}

\begin{table*}[ht]
\centering
\small
\begin{tabular}{@{\extracolsep{\fill}}cccccc@{}}
\toprule
\multicolumn{6}{c}{\textbf{Mention Ambiguity (dim=20)}} \\
\midrule
\textbf{Layer} & \textbf{GPT-2} & \textbf{Llama-2 7B} & \textbf{Llama-2 13B} & \textbf{Llama-3 8B} & \textbf{Mistral 7B} \\
\midrule
0 & 0.535 & 0.381 & 0.385 & 0.384 & 0.381 \\
1 & 0.767 & 0.808 & 0.819 & 0.828 & 0.826 \\
2 & 0.769 & 0.817 & 0.816 & 0.836 & 0.821 \\
3 & 0.793 & 0.819 & 0.843 & 0.844 & 0.832 \\
4 & 0.792 & 0.849 & 0.864 & 0.869 & 0.836 \\
5 & 0.796 & 0.848 & 0.870 & 0.870 & 0.841 \\
6 & 0.797 & 0.855 & 0.873 & 0.875 & 0.873 \\
7 & 0.798 & 0.859 & 0.867 & 0.875 & 0.867 \\
8 & 0.793 & 0.868 & 0.881 & 0.878 & 0.879 \\
9 & 0.795 & 0.869 & 0.882 & 0.874 & 0.883 \\
10 & 0.795 & 0.868 & 0.888 & 0.876 & 0.884 \\
11 & 0.795 & 0.870 & 0.887 & 0.873 & 0.880 \\
12 & 0.801 & 0.866 & 0.881 & 0.876 & 0.880 \\
13 & -- & 0.867 & 0.870 & 0.869 & 0.886 \\
14 & -- & 0.854 & 0.869 & 0.866 & 0.876 \\
15 & -- & 0.850 & 0.871 & 0.866 & 0.869 \\
16 & -- & 0.844 & 0.873 & 0.851 & 0.862 \\
17 & -- & 0.841 & 0.865 & 0.850 & 0.857 \\
18 & -- & 0.844 & 0.865 & 0.848 & 0.851 \\
19 & -- & 0.846 & 0.865 & 0.848 & 0.848 \\
20 & -- & 0.844 & 0.859 & 0.840 & 0.851 \\
21 & -- & 0.850 & 0.859 & 0.839 & 0.850 \\
22 & -- & 0.852 & 0.867 & 0.846 & 0.854 \\
23 & -- & 0.850 & 0.871 & 0.841 & 0.855 \\
24 & -- & 0.860 & 0.870 & 0.839 & 0.855 \\
25 & -- & 0.857 & 0.872 & 0.840 & 0.853 \\
26 & -- & 0.862 & 0.874 & 0.840 & 0.858 \\
27 & -- & 0.866 & 0.875 & 0.846 & 0.863 \\
28 & -- & 0.871 & 0.874 & 0.849 & 0.860 \\
29 & -- & 0.868 & 0.874 & 0.850 & 0.861 \\
30 & -- & 0.871 & 0.872 & 0.855 & 0.868 \\
31 & -- & 0.873 & 0.871 & 0.860 & 0.866 \\
32 & -- & 0.877 & 0.871 & 0.878 & 0.877 \\
33 & -- & -- & 0.869 & -- & -- \\
34 & -- & -- & 0.876 & -- & -- \\
35 & -- & -- & 0.876 & -- & -- \\
36 & -- & -- & 0.877 & -- & -- \\
37 & -- & -- & 0.881 & -- & -- \\
38 & -- & -- & 0.881 & -- & -- \\
39 & -- & -- & 0.881 & -- & -- \\
40 & -- & -- & 0.877 & -- & -- \\
\bottomrule
\end{tabular}
\caption{AUC scores measuring the area under the curve where the x-axis represents mention ambiguity and the y-axis represents F1 scores derived from Purity and IP metrics across different model layers.
All embeddings are reduced to 20 dimensions. }
\label{tab:auc_mention_amb}
\end{table*}

\begin{table*}[ht]
\centering
\small
\begin{tabular}{@{\extracolsep{\fill}}cccccc@{}}

\toprule
\multicolumn{6}{c}{\textbf{Mention Variability (dim=20)}} \\
\midrule

\textbf{Layer} & \textbf{GPT-2} & \textbf{Llama-2 7B} & \textbf{Llama-2 13B} & \textbf{Llama-3 8B} & \textbf{Mistral 7B} \\
\midrule
0 & 0.377 & 0.290 & 0.290 & 0.441 & 0.305 \\
1 & 0.574 & 0.473 & 0.506 & 0.737 & 0.693 \\
2 & 0.589 & 0.695 & 0.748 & 0.805 & 0.780 \\
3 & 0.630 & 0.779 & 0.806 & 0.817 & 0.803 \\
4 & 0.644 & 0.795 & 0.825 & 0.822 & 0.826 \\
5 & 0.654 & 0.812 & 0.825 & 0.833 & 0.819 \\
6 & 0.663 & 0.815 & 0.839 & 0.825 & 0.826 \\
7 & 0.650 & 0.818 & 0.832 & 0.824 & 0.821 \\
8 & 0.658 & 0.817 & 0.831 & 0.831 & 0.816 \\
9 & 0.648 & 0.797 & 0.822 & 0.823 & 0.810 \\
10 & 0.646 & 0.794 & 0.810 & 0.824 & 0.802 \\
11 & 0.648 & 0.797 & 0.803 & 0.824 & 0.804 \\
12 & 0.645 & 0.791 & 0.794 & 0.822 & 0.802 \\
13 & -- & 0.784 & 0.782 & 0.811 & 0.795 \\
14 & -- & 0.781 & 0.771 & 0.811 & 0.788 \\
15 & -- & 0.766 & 0.764 & 0.799 & 0.783 \\
16 & -- & 0.761 & 0.773 & 0.787 & 0.764 \\
17 & -- & 0.748 & 0.758 & 0.776 & 0.754 \\
18 & -- & 0.729 & 0.753 & 0.768 & 0.751 \\
19 & -- & 0.723 & 0.743 & 0.756 & 0.737 \\
20 & -- & 0.711 & 0.722 & 0.742 & 0.725 \\
21 & -- & 0.701 & 0.721 & 0.725 & 0.715 \\
22 & -- & 0.692 & 0.710 & 0.730 & 0.704 \\
23 & -- & 0.693 & 0.701 & 0.718 & 0.692 \\
24 & -- & 0.686 & 0.693 & 0.717 & 0.690 \\
25 & -- & 0.682 & 0.691 & 0.723 & 0.689 \\
26 & -- & 0.673 & 0.688 & 0.722 & 0.680 \\
27 & -- & 0.670 & 0.681 & 0.727 & 0.678 \\
28 & -- & 0.665 & 0.680 & 0.724 & 0.682 \\
29 & -- & 0.657 & 0.675 & 0.726 & 0.681 \\
30 & -- & 0.649 & 0.674 & 0.737 & 0.685 \\
31 & -- & 0.647 & 0.672 & 0.738 & 0.678 \\
32 & -- & 0.635 & 0.668 & 0.665 & 0.662 \\
33 & -- & -- & 0.663 & -- & -- \\
34 & -- & -- & 0.657 & -- & -- \\
35 & -- & -- & 0.655 & -- & -- \\
36 & -- & -- & 0.651 & -- & -- \\
37 & -- & -- & 0.648 & -- & -- \\
38 & -- & -- & 0.653 & -- & -- \\
39 & -- & -- & 0.660 & -- & -- \\
40 & -- & -- & 0.655 & -- & -- \\
\bottomrule
\end{tabular}
\caption{AUC scores measuring the area under the curve where the x-axis represents mention variability and the y-axis represents F1 scores derived from Purity and IP metrics across different model layers.
All embeddings are reduced to 20 dimensions.}
\label{tab:auc_mention_vari}
\end{table*}

\subsubsection{Model Comparison Results}
\label{sec:appendix_mention_variability_models} 
\begin{figure*}[t]
\centering
\includegraphics[width=16cm]{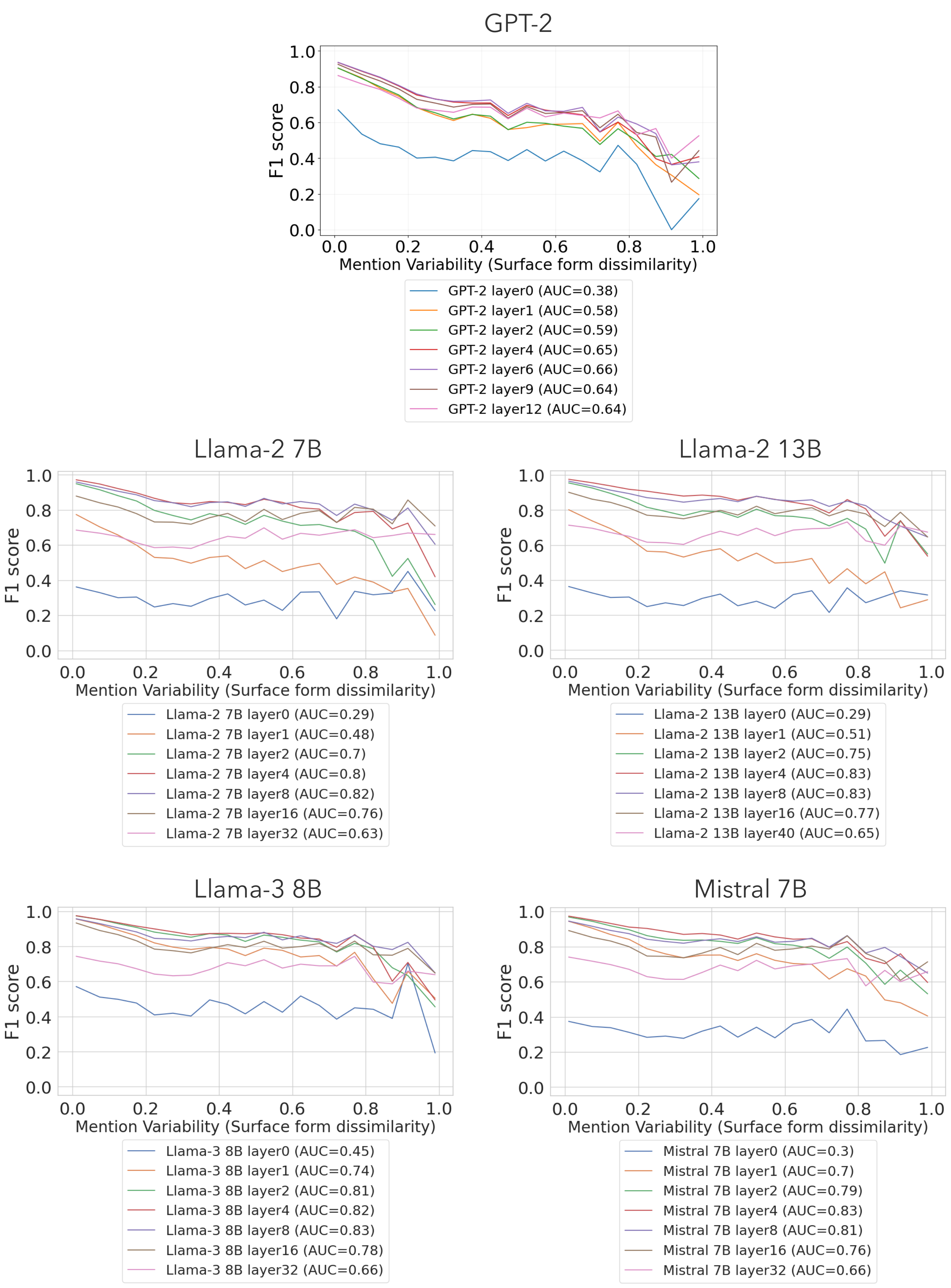}
\caption{
Results for all models: mention variability and the separability of entity representations (F1 score)
}
\label{fig:mention_variability_result_summary_all_model}
\end{figure*}

Table~\ref{tab:baseline_auc} shows the AUC of mention variability and F1 score for the baseline. 
Table~\ref{tab:auc_mention_vari} and Figure \ref{fig:mention_variability_result_summary_all_model} show the analysis results from layer 0 to the final layer for mention variability using GPT-2, Llama-2 7B, Llama-2 13B, Llama-3 8B, and Mistral 7B.

\subsubsection{Results with Average Subword Token Embeddings}
\label{subsec:appendix_mention_vari_ave}
Figure~\ref{fig:result_summary_ave} shows the results when using average embeddings of subword tokens as the target tokens for analysis.
The cross-model trends remain consistent with those observed in Figure~\ref{fig:result_summary} (\S~\ref{sec:result}).
Notably, however, Llama-2 7B at layer 0 (bottom right in Figure~\ref{fig:result_summary_ave}) achieves higher F1 scores compared to the last token embedding results in Figure~\ref{fig:result_summary}.
This improvement likely stems from the averaging process enabling greater reliance on surface-level character information for entity identification.

\subsection{Qualitative Analysis of Output Inconsistency}
\label{subsec:appendix_output_inconsistency}
We conducted a qualitative analysis to investigate the impact of mention variability on model outputs. Interestingly, even when outputs were inconsistent across different surface forms, the predicted years in patterns like ``[X] was founded in [Y]'' remained temporally proximate. 
For example:
\begin{itemize}
   \item ``Red Crescent Society was founded in 1934'' vs. ``Red Crescent was founded in 1919'' (15-year difference)
   \item ``Holy Cross Brothers was founded in 1847'' vs. ``Brothers of Holy Cross was founded in 1845'' (2-year difference)
\end{itemize}
This suggests that while models struggle with exact consistency across mention variants, they maintain approximate temporal coherence, capturing the general historical period rather than memorizing exact dates.

\section{Supplementary Experiment: Adjusted Rand Index for Entity Identification}
\label{sec:appendix_ari}
In addition to the Purity and Inverse Purity (IP) metrics presented in the main body of this paper, we conducted a supplementary experiment using the Adjusted Rand Index (ARI)~\citep{ARI} to evaluate the extent to which our models' internal representations facilitate entity identification. The Adjusted Rand Index is a measure of the similarity between two data clusterings, accounting for chance. In this context, it serves as an alternative approach to assessing the quality of entity identification, complementing our primary metrics.

\subsection{Experimental Setup}
The experiment was designed to quantify how well mentions of the same entity cluster together in the embedding space. 
We used the same dataset and entity representation definitions as described in Section~\ref{sec:Experiments_setup}.
Cluster formation followed the same methodology as detailed in Section~\ref{subsec:Purity_IP_F1}, primarily forming clusters based on the entity of the nearest centroid to a given mention embedding.
The Adjusted Rand Index was then computed between these generated clusters and the gold-standard entity labels.
The Adjusted Rand Index (ARI) is calculated as:
\begin{align}
\text{ARI} = \frac{\text{RI} - E[\text{RI}]}{\max(\text{RI}) - E[\text{RI}]}
\end{align}
where $\text{RI}$ is the Rand Index, $E[\text{RI}]$ is the expected Rand Index under random partitioning, and $\max(\text{RI})$ is the maximum possible Rand Index (which is 1). The Rand Index ($\text{RI}$) itself measures the agreement between two clusterings by considering all pairs of samples and counting pairs that are assigned in the same or different clusters in both the predicted and true clusterings. ARI corrects for chance agreement, ensuring that random labelings have an ARI close to 0.0, and perfect agreement results in an ARI of 1.0.

\subsection{Results}
\begin{figure*}[t]
\centering
\includegraphics[width=\linewidth]{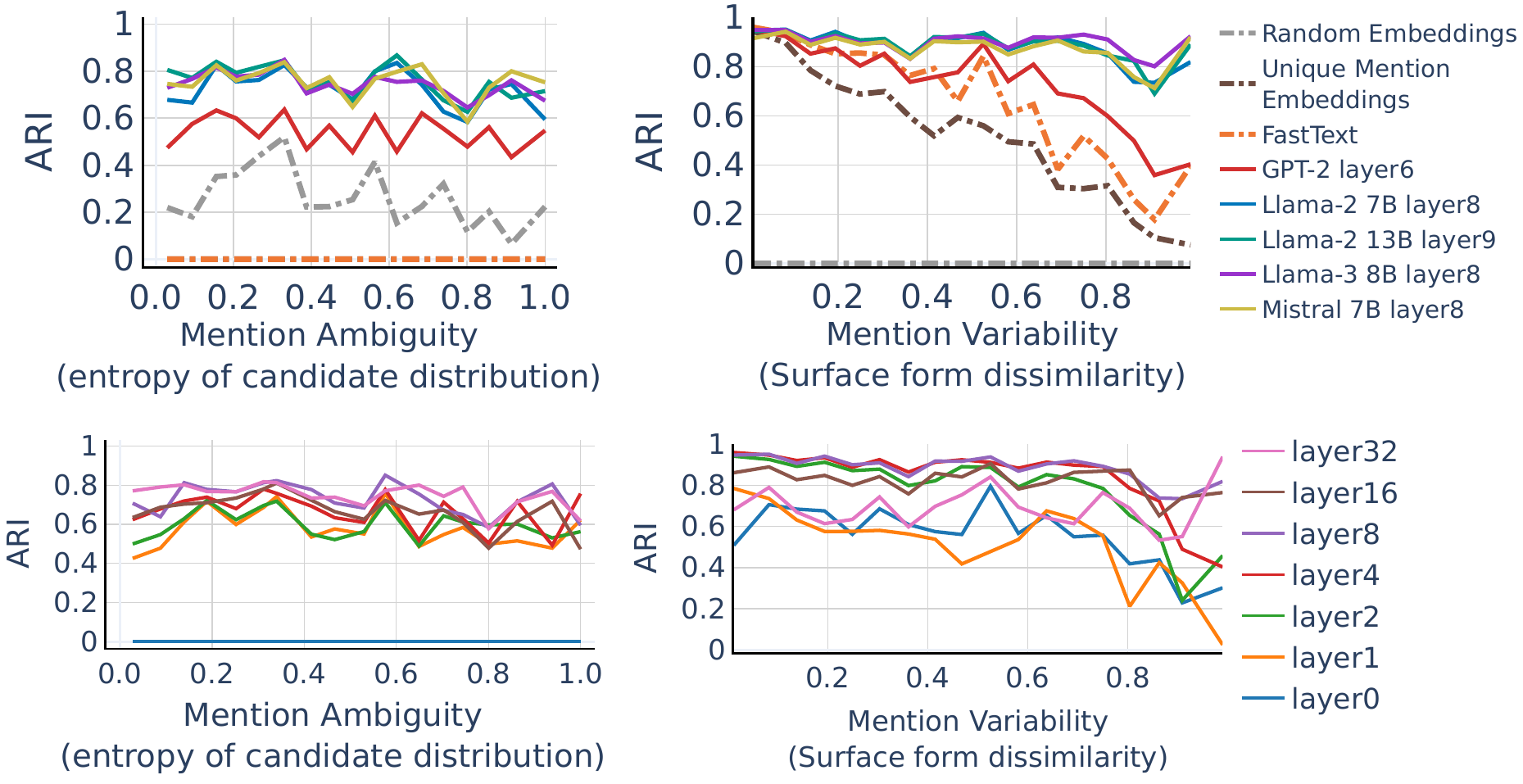}
\caption{
ARI (Adjusted Rand Index) of entity representations for each difficulty~(left: mention ambiguity, right: mention variability). 
ARI scores are averaged within each bin on the x-axis. 
Top: Results across all models, using the layer with highest AUC for LM scores. 
Bottom: Layer-wise analysis for Llama-2 7B. 
}
\label{fig:adjusted_rand_index_results}
\end{figure*}

The results of this supplementary experiment are presented in Figure \ref{fig:adjusted_rand_index_results}. Consistent with the trends observed using Purity and IP, the Adjusted Rand Index scores reveal a largely similar relative ranking among the models. Specifically, models that performed well on Purity and IP generally also achieved higher Adjusted Rand Index scores. This consistency further validates our findings regarding the models' ability to distinguish between different entities based on their contextual embeddings.

\section{Computational Resources}
For this experiment, we utilized four NVIDIA RTX 6000 Ada graphics cards.

\section{Usage of AI assistants}
We used an AI Assistant (Claude) for writing this paper and developing source code for the experiments. However, its use was limited to code completion, translation, text refinement, and table creation, while the content and ideas are solely those of the authors.

\end{document}